\documentclass{article}

\usepackage{microtype}
\usepackage{graphicx}
\usepackage{subcaption}
\usepackage{booktabs}
\usepackage{multirow}
\usepackage{xspace}

\usepackage{hyperref}

\usepackage[accepted]{icml2026}
\usepackage{amsmath}
\usepackage{amssymb}
\usepackage{mathtools}
\usepackage{amsthm}
\usepackage[capitalize,noabbrev]{cleveref}

\theoremstyle{plain}

\theoremstyle{definition}

\theoremstyle{remark}

\usepackage[textsize=tiny]{todonotes}

\newcommand{\ours}{Hibiki-Zero\xspace}

\newcommand{\ensuremathmode}[1]{%
  \ifmmode
    #1% Already in math mode, use as is
  \else
    $#1$% Not in math mode, wrap with dollar signs
  \fi
}
\newcommand{\softmax}{\mathrm{softmax}}

\DeclareMathOperator*{\E}{\mathbb{E}}
\newcommand{\proba}[1]{\mathbb{P}\left[#1\right]}

\begin{document}

\twocolumn[
\icmltitle{Simultaneous Speech-to-Speech Translation Without Aligned Data}

\icmlsetsymbol{equal}{*}

\begin{icmlauthorlist}
\icmlauthor{Tom Labiausse}{kyutai}
\icmlauthor{Romain Fabre}{kyutai}
\icmlauthor{Yannick Est\`eve}{lia}
\icmlauthor{Alexandre D\'efossez}{kyutai,gradium}
\icmlauthor{Neil Zeghidour}{gradium}
\end{icmlauthorlist}

\icmlaffiliation{kyutai}{Kyutai, Paris, France}
\icmlaffiliation{gradium}{Gradium, Paris, France}
\icmlaffiliation{lia}{LIA, University of Avignon, France}

\icmlcorrespondingauthor{\\ \ours{}}{hibiki@kyutai.org}

\vskip 0.3in
]

\printAffiliationsAndNotice{}  % no special notice (required even if empty)

\begin{abstract}
Simultaneous speech translation requires translating source speech into a target language in real-time while handling non-monotonic word dependencies. Traditional approaches rely on supervised training with word-level aligned data, which is difficult to collect at scale and thus depends on synthetic alignments using language-specific heuristics that are suboptimal. We propose \ours{}, which eliminates the need for word-level alignments entirely. This fundamentally simplifies the training pipeline and enables seamless scaling to diverse languages with varying grammatical structures, removing the bottleneck of designing language-specific alignment heuristics. We first train on sentence-level aligned data to learn speech translation at high latency, then apply a novel reinforcement learning strategy using GRPO to optimize latency while preserving translation quality. \ours{} achieves state-of-the-art performance in translation accuracy, latency, voice transfer, and naturalness across four X-to-English tasks.  Moreover, we demonstrate that our model can be adapted to support a new input language with less than 1000h of speech. We provide examples, model weights, inference code\footnote{\href{https://github.com/kyutai-labs/hibiki-zero}{github.com/kyutai-labs/hibiki-zero}} and we release a benchmark containing 45h of multilingual data for speech translation evaluation.\footnote{\href{https://huggingface.co/collections/kyutai/hibiki-zero}{huggingface.co/collections/kyutai/hibiki-zero}}
\end{abstract}

\section{Introduction}
\label{introduction}

\looseness=-1
We introduce \ours{}, a system for simultaneous and expressive speech-to-speech (S2ST) and speech-to-text (S2TT) translation that does not require aligned data for training. Unlike offline speech translation systems that access the full source utterance before translating, simultaneous translation must produce output incrementally while maintaining both translation accuracy and speech naturalness. This requires learning a fine-grained translation policy that determines when to listen and when to speak. The most straightforward approach to learning such a policy is through supervised training on aligned data. However, human interpretation data with word-level alignments is virtually non-existent, forcing state-of-the-art systems to rely on synthetic data with automatic alignments~\cite{hibiki}. These automatic alignments are inherently limited, as they depend on hand-crafted heuristics rather than being learned from data.

\looseness=-1

\ours{} is a decoder-only model that synchronously receives source speech and generates translated speech leveraging a multistream architecture originally introduced by~\citet{moshi}. Unlike Hibiki~\cite{hibiki}, \ours{} is not trained with supervised learning on synthetic interpretation data but rather casts joint optimization of translation quality and latency as a reinforcement learning (RL) problem. While we still require a base model before the RL phase, it is trained using sentence-level aligned data which can be more easily constructed independently of the language compared to word-level aligned data. During RL, we exploit the sentence-level aspect of our data to design a simple reward system based on BLEU score~\cite{bleu} only. To achieve this, we compute rewards at multiple intermediate instants during the translation of an input speech utterance by leveraging the simultaneous text translation also produced by our model. Using these \textit{process rewards}, we obtain fine-grained local \textit{advantages} across multiple translations from the same input. We then adapt GRPO~\cite{grpo} to our multistream architecture, using these advantages to optimize the model.

\looseness=-1
In a multilingual-to-English translation task, \ours{} outperforms previous state-of-the-art work in translation quality, latency, speaker identity preservation, and speech naturalness. We also retain all the benefits of multistream modeling such as batching and real-time inference on GPU while removing the necessity to build interpretation-like training data thus considerably simplifying the development of such models. We even demonstrate that \ours{} can adapt to a new input language with less than 1000h of training data marking an important step to make high quality speech translation (ST) available in more languages.

\paragraph{Conflict of interest disclosure.} Authors N.Z. and A.D. are employed by Gradium, which develops one of the three TTS models used to create the long-form evaluation dataset.
\section{Related Work}
\label{related-work}

\subsection{Simultaneous end-to-end speech translation}
\looseness=-1

While speech translation was initially performed using cascaded systems combining automatic speech recognition (ASR), machine translation (MT) and text-to-speech synthesis (TTS)~\cite{wahlster2000verbmobil, nakamura2006atr}, it recently evolved in fully end-to-end systems~\cite{jia19_translatotron,lee-etal-2022-direct,jia22-translatotron2,audiopalm} reducing error propagation and enabling transfer of non-linguistic information such as the speaker voice identity or prosody to the generated speech. At first trained with auxiliary text of phoneme translation tasks~\cite{jia22-translatotron2, streamspeech}, most recent works~\cite{seamless,hibiki,seed_liveinterpret_2, google_s2st} train directly on simultaneous S2TT and S2ST tasks so they can use the predicted text translation as a scaffolding for speech generation at inference time. Among direct ST training methods, those who achieve better speech naturalness are duplex audio systems that require to build a simultaneous ST dataset. They either rely on a synthetic data generation pipeline which includes a fine word-level text-to-translation alignment method~\cite{hibiki, google_s2st} or use a text LLM to split text into semantic chunks (a few words) that are individually translated thus providing chunk-level translation alignment~\cite{seed_liveinterpret_2} before collecting human-annotated interpretation data for finetuning purposes. \ours{} removes most of the complexity from synthetic data generation as it only requires sentence-level translation alignment easily obtained from punctuation. Thanks to an efficient RL process, it is then possible to reduce the translation latency of the model so it achieves state-of-the-art quality/latency trade-off in multiple input languages.

\subsection{Self-improvement of real-time translation systems}
\looseness=-1

RL methods to improve simultaneous translation systems were first explored in the context of text translation. Some works used preference-based approaches~\cite{simulpl} with preferences established in the context of simultaneous ST by prompting a text LLM while others applied online reinforcement procedures~\cite{simulpl, seqpo_simt} with sequence-level rewards as a combination of translation quality and latency metrics. Because they lack sub-sentence granularity in their preference or reward signals, it is difficult for these methods to find an appropriate balance between translation quality and latency during the RL process. More recently, Seed LiveInterpret 2.0~\cite{seed_liveinterpret_2} applied PPO~\cite{ppo} with a combination of intermediate evaluations of the generated sequences (\textit{process rewards}) and overall evaluation of the translation (\textit{outcome rewards}). Starting from a base supervised ST model trained with chunk-level alignment and finetuned on high-quality human interpretation data, they managed to strictly improve the quality/latency trade-off through RL. However, they combine multiple quality and latency measures to compute their reward thus introducing hyperparameters that are difficult to tune in practice. Because of complex interactions between the different reward components, they encountered training instability, reward hacking and had to rely on two different stages of RL training, using only outcome rewards at first before adding process rewards. On the other hand, \ours{} uses a single and straightforward reward system that unifies the translation quality and latency criteria into a single reward derived from BLEU~\cite{bleu} scores only. Reinforcement learning is performed using GRPO~\cite{grpo} without KL regularization as previously done by~\citet{magistral} to reduce memory requirements during training. Most importantly, it does not rely on any human interpretation or annotated data to finetune the model before reinforcement. On multilingual simultaneous ST tasks, \ours{} achieves state-of-the-art translation quality, translation latency, naturalness and speaker identity preservation. \ours{} is even able to adapt to a new input language after a light finetuning.
\section{Method}
\label{method}

We consider an utterance in a source language represented as
a monophonic waveform $X \in\mathbb{R}^{f_s \cdot d}$, sampled at a frame rate $f_s = 24\,\mathrm{kHz}$, of duration $d$.
Similarly, its translation is given in a target language,
denoted $Y \in \mathbb{R}^{f_s \cdot d}$. We assume $X$ is padded to ensure both have the same duration. Our objective is to model $\proba{Y | X}$. Contrary to \citet{hibiki}, we do not constrain the modeling of $Y$ knowing $X$ to be entirely causal in our training data. Thanks to the diversity of causality and latency arrangements in the dataset, it is still possible to learn a base translation model. Its behavior is then adjusted by an online reinforcement learning strategy that rewards correct and simultaneous translations.

\subsection{Modeling}
\label{sec:modeling}

We build on the framework introduced by \citet{moshi} for the joint modeling of
multiple sequences of tokens and used by \citet{hibiki} to perform simultaneous S2TT and S2ST with high fidelity.

\subsubsection{Neural audio codec}
\label{sec:neural_audio_codec}

We use the pre-trained causal and streaming Mimi codec~\citep{moshi} to encode $X$ and $Y$ into low framerate sequences of discrete tokens.
Mimi consists of an encoder and decoder from and to the waveform domain, and of an information bottleneck using Residual Vector Quantization (RVQ)~\cite{soundstream}.

For language modeling, we are interested in the discrete
indices of codebook entries which Mimi latents are projected to. We denote those $(A_{t, q}) \in \{1, \ldots, N_a\}^{f_r \cdot d \times Q}$ where $f_r = 12.5 \,\text{Hz}$ is the codec framerate, $Q$ is the number of audio residual quantization levels varying up to 32 and $N_a$ the codebooks size. Following~\citet{zhang2024speechtokenizer,moshi}, the output
of the first quantization level is trained to replicate semantic information
obtained from a WavLM self-supervised audio model~\citep{wavlm}. We refer to $A_{t, 1}$ as \emph{semantic} tokens, and $A_{t, q \geq 2}$ as \emph{acoustic} tokens with the latter arranged in a coarse to fine manner. When encoding audio, we keep only the first $Q=16$ acoustic levels as they are sufficient to ensure high quality speech reconstruction through Mimi's decoder.

\begin{figure}[t]
    \centering
    \includegraphics[width=0.8\columnwidth]{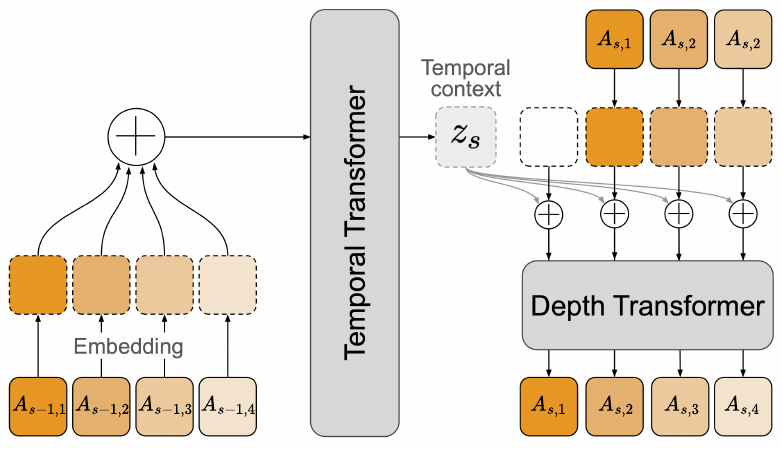}
    \caption{\textbf{Architecture of the RQ-Transformer.} Figure adapted from ~\citet{moshi}.}
    \label{fig:rq-transformer}
\end{figure}

\begin{figure}[t]
    \centering
    \includegraphics[width=\columnwidth]{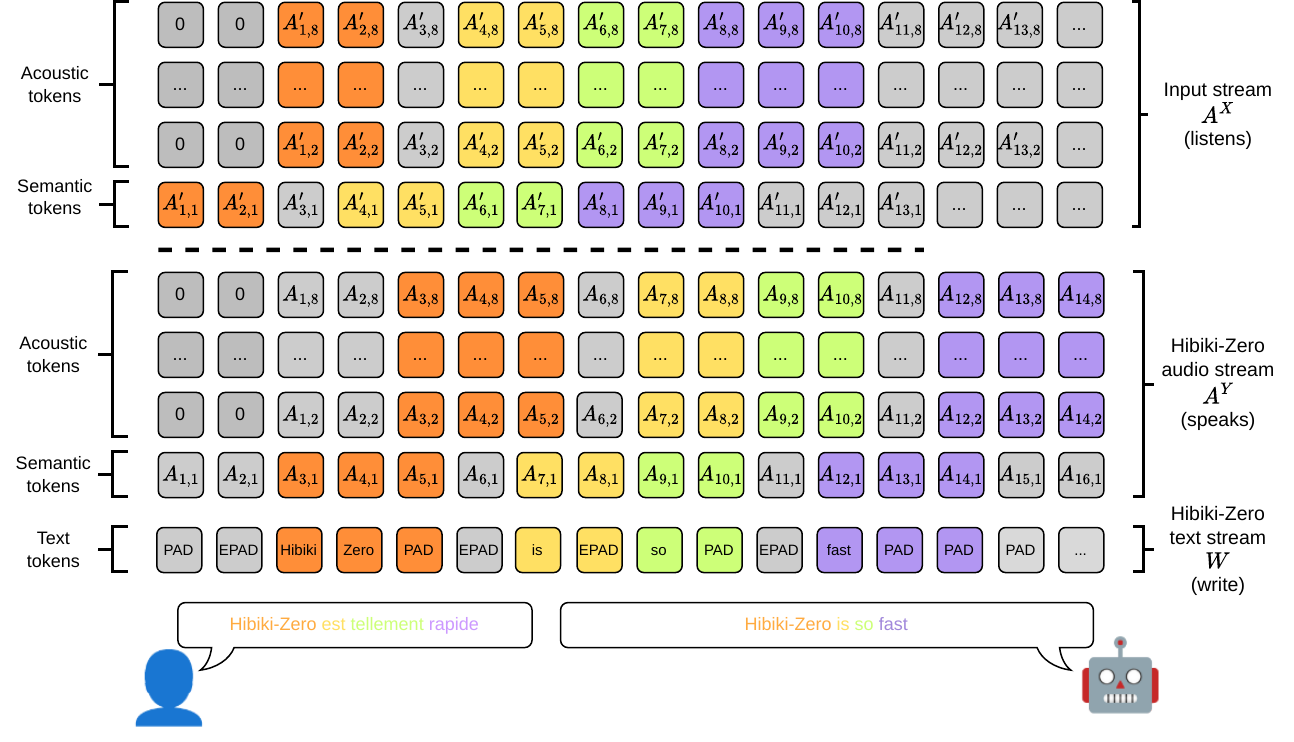}
    \caption{\textbf{Joint sequence modeling.} From the input stream, \ours{} predicts its Inner Monologue text stream, semantic and acoustic tokens. Figure adapted from ~\citet{hibiki}.}
    \label{fig:joint-modeling}
\end{figure}

\subsubsection{Joint modeling of discrete audio tokens}
\label{sec:joint_modeling}

Following \citet{uniaudio,hibiki}, we leverage a RQ-Transformer~\citep{rq-transformer} as shown in Figure~\ref{fig:rq-transformer} to model $(A_{t, q})$ both over the time $t$ and quantizer $q$ axes as audio streams cannot be reasonably merged into a single discrete sequence.
It consists in a large \emph{Temporal} Transformer~\citep{attentionvaswani} of latent dimension $D$, operating at the same framerate $f_r$ as the codec, and being fed all the tokens generated so far, e.g.
for all $t \leq f_r \cdot d$, 
 \begin{equation}
 \label{eq:temp_transformer}
Z_t = \mathrm{Temp}(A_0, \ldots, A_{t - 1}) \in \mathbb{R}^{D}.
\end{equation}
$A_0$ is defined as a deterministic token indicating the start of the generation.
Then, a smaller scale \emph{Depth} Transformer models auto-regressively
the tokens $A_{t, 1}, \ldots, A_{t, Q}$ over the quantizer axis, e.g. for all $t \leq f_r \cdot d$ and $q \leq Q$,
\begin{equation}
\label{eq:dep_transformer}
    l_{t, q} = \mathrm{Dep}(Z_t, A_{t, 0}, \ldots, A_{t, q - 1}) \in \mathbb{R}^{N_a},
\end{equation}
with $A_{t, 0}$ also a special token, and with the goal of having,
\begin{equation*}
    \softmax(l_{t, q}) \approx \proba{A_{t, q} | A_{0}, \ldots, A_{t-1}, A_{t, 0}, \ldots A_{t, q - 1}}
\end{equation*}

Following \citep{musicgen,moshi}, we introduce an acoustic delay shifting acoustic tokens of 2 time steps in the future compared to the semantic stream. The streams are realigned before decoding the audio with the codec. As this delay is always applied, we don't introduce new notations for readability and refer to $(A_{t,q})$ directly.

\subsubsection{Translation as multistream modeling}
\label{sec:multistream_modeling}

Using the RQ-Transformer given by Eq.~\eqref{eq:temp_transformer} and \eqref{eq:dep_transformer} to jointly model multiple discrete streams of tokens, we can perform the task of joint simultaneous S2TT and S2ST as illustrated in Figure~\ref{fig:joint-modeling}.
Following~\cite{moshi}, we use an \textit{Inner Monologue} to improve stability of the generated audio content. We introduce a stream of padded text tokens $(W_t) \in \{1, \ldots, N_W\}^{f_r \cdot d}$ whose content is the word-level aligned text transcription of the audio modeled in $A^Y$. This text stream is concatenated with the audio tokens $A^Y$ along the $q$-axis such that it comes before the semantic level. The resulting multistream of target tokens are themselves concatenated with the source tokens $A^X$ along the $q$-axis.
At inference time, predictions of tokens $A^X$ are skipped and actual tokens of the input audio are forced instead.

\subsubsection{Architectural details}
\label{sec:arch_details}

At time-step $t$, tokens from the previous step, e.g. $A^X_{t-1}$,
$A^Y_{t-1}$, and $W_{t-1}$, are fed into dedicated embedding tables and contributions are summed with a BOS token used for the first time step $t = 1$.
The RQ-Transformer uses standard Transformer layers~\citep{attentionvaswani}, with gated SiLU activation~\citep{shazeer2020glu,hendrycks2016gaussian}.
A linear layer maps output $Z_t$ of the \emph{Temporal} Transformer to logits for the text token $W_t$.
The \emph{Depth} Transformer then operates for $Q$ steps to estimate the logits for the output stream and for $Q$ additional steps for the input stream.
Each depth step $q$ takes as input $Z_t$ summed with a learned embedding of the previous audio token $A_{t, q - 1}$, or $W_t$ for $q = 1$.
We provide architectural hyper-parameters in Section~\ref{sec:arch_hyper_params}.

%%% SECTION %%%
\subsection{Coarse alignment of speech translation data}
\label{sec:synth_st_data}

We have assumed training pairs $(X, Y)$ to not be entirely causal at the interpretation level. We now detail the specific method used to build such coarse translation alignments.

\subsubsection{Sentence-level alignment}
\label{sec:sent_level_align}

We start from an unaligned speech translation pair $(X, Y)$ which only verifies a sentence mapping constraint meaning that both $X$ and $Y$ contain the same number of sentences and such that the $i^{th}$ sentence in $Y$ is a translation of the $i^{th}$ sentence in $X$. Inspired by \citet{hibiki}, we rely on the insertion of artificial silence in $Y$ to delay its content with respect to $X$.  For each sentence of index $i$, we introduce silence in $Y$ to shift its $i^{th}$ sentence by an amount $\delta_i$ after the start of the $i^{th}$ sentence in $X$ where $\delta_i \sim \mathcal{U}(0,\delta \times d_{i})$ is sampled independently for each sentence, $d_{i}$ is the duration of the $i^{th}$ sentence in $X$ and $\delta \in [0,1]$ is an hyperparameter. Then, using punctuation characters such as commas or colons in a precomputed transcript of $Y$, we insert silences whose durations follow $\mathcal{U}(0,\mu)$ at the corresponding instants in $Y$ with $\mu$ being a hyperparameter.

\subsubsection{Natural pauses TTS}
\label{sec:nat_pauses_tts}

Using the method described in Section \ref{sec:sent_level_align}, we might break the natural flow of speech by inserting silence on punctuations which is also subject to imprecisions of the transcript timestamps. Following \citet{dsm}, we train a TTS with synced audio and text streams as output, providing a control on the emission timestamp of each word to synthesize. Moreover, we train the TTS to perform voice transfer from a short audio conditioning of maximum 10 seconds. We can then generate an audio $Y^+$ using the original transcript of $Y$ and naturally insert the pauses described in \ref{sec:sent_level_align} while conditioned on the speaker from $X$. This results in new training pairs ($X$, $Y^+$) where targets contain smoother transitions between speech and silences than $Y$.

%%% SECTION %%%
\subsection{Translation policy reinforcement}
\label{sec:rl_method}
\begin{figure}[t]
    \centering
    \includegraphics[width=\columnwidth, trim={0 0.03cm 5cm 0}, clip]{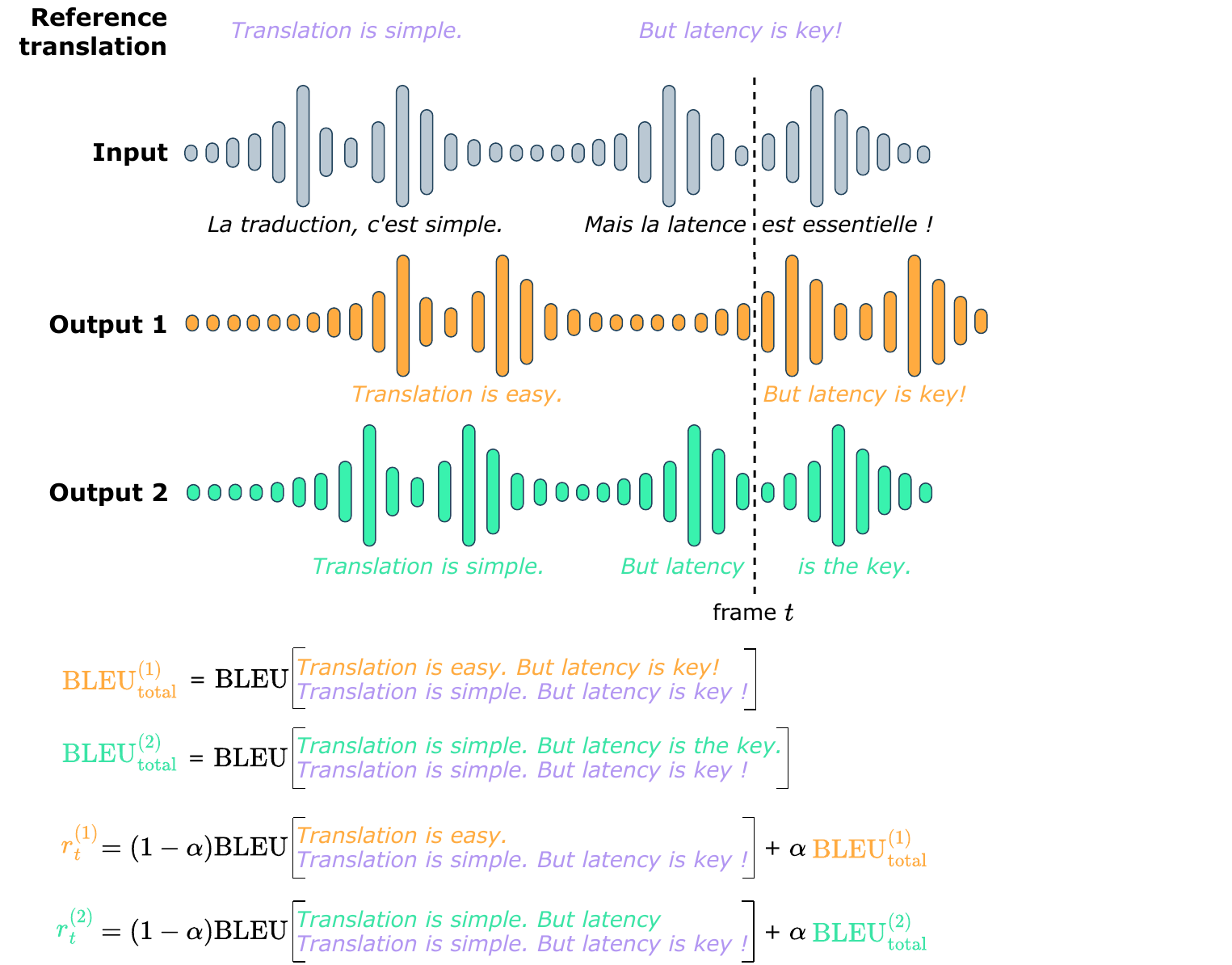}
    \caption{\textbf{Process rewards method based on BLEU score.} We introduce intermediate BLEU score computed on the text output of the model before a given frame $t$ and using the ground-truth translation of the corresponding input sentences processed so far. We combine it with the total output BLEU score using $\alpha \in [0,1]$.}
    \label{fig:rewards_pattern}
\end{figure}

Assuming that we dispose of a simultaneous translation model as presented in Section~\ref{sec:modeling}, we now introduce a reinforcement learning procedure using process rewards based on BLEU scores to improve the translation policy of the model as illustrated in Figure~\ref{fig:rewards_pattern}. We adapt GRPO from \citet{grpo} to be our RL algorithm. We denote by $\pi_{\theta}$ the translation model to optimize and $\pi_{\theta_{\mathrm{old}}}$ an older version of it acting as a regularizer. Given an input speech utterance $X$ with a known sentence-level text translation $y$, we use $\pi_{\theta_{\mathrm{old}}}$ to generate $G$ different speech translations $(Y_i)_{1 \le i \le G}$, each of duration $T \times f_r$ seconds where $f_r$ is the model frame rate and $T$ a fixed number of frames.

\subsubsection{Process rewards}
\label{sec:process_rewards}

\looseness=-1
Let $n$ be the number of sentences in $X$ and $(t_i)_{0 \le i \le n}$ the frame indexes such that the sentence of index $i$ in $X$ starts at frame $t_i$ and ends at frame $t_{i+1}$. We introduce $S(t)$ as the sentence index at frame $t \ge t_0$ in $X$ i.e. $S(t)=i$ for $t_i \le t \le t_{i+1}$ and $S(t)=n-1$ for $t > t_n$. For a frame index $t \le T$, we denote $y_t$ as the text concatenation of translated input sentences until the one of index $S(t)$ included. Given a generation $i$, we define $\hat{y}^{(i)}_{t}$ as the partial text transcript until frame $t$ given by the model's output text stream. We now introduce the hyperparameter $\alpha \in [0,1]$ and define the process reward for generation $i$ at frame $t$ as:
\begin{equation}
\label{eq:process_reward}
    r^{(i)}_{t} = (1 - \alpha) \mathrm{BLEU} \big( \hat{y}^{(i)}_{t}, y_{t} \big) + \alpha \mathrm{BLEU} \big( \hat{y}^{(i)}_{T}, y_{T} \big)
\end{equation}

\subsubsection{Optimization objective}
\label{sec:optim_objective}
Using the modeling of $X$ and $(Y_i)_{1 \le i \le G}$ as tokens $A^X$ and $A^{Y_{1}}$, ..., $A^{Y_{G}}$, we define the probability ratios between $\pi_{\theta}$ and $\pi_{\theta_{\mathrm{old}}}$ for each output $i$, codebook index $q \le Q$ and frame index $t \le T$ as:
\begin{equation}
\label{eq:log_prob}
    p^{(i)}_{q,t} = \frac{\pi_{\theta} \Big( A^{Y_{i}}_{q,t}|A^X_{\le t},A^{Y_{i}}_{<t},A^{Y_{i}}_{<q,t} \Big) }{\pi_{\theta_{\mathrm{old}}} \Big( A^{Y_{i}}_{q,t}|A^X_{\le t},A^{Y_{i}}_{<t},A^{Y_{i}}_{<q,t} \Big) }
\end{equation}

Given a set of frame indexes $t'_1 < t'_2 < ...< t'_s$, we compute process rewards as defined in Section~\ref{sec:process_rewards} for each output, namely $r^{(i)}_{t'_{j}}$ for $i \le G$ and $j \le s$. We then normalize rewards per frame index across group elements:
\begin{equation}
\label{eq:norm_reward}
\bar{r}^{(i)}_{t'_{j}} = \frac{r^{(i)}_{t'_{j}} - \underset{k \le G}{\mathrm{mean}} \Big[ r^{(k)}_{t'_{j}} \Big]}{\underset{k \le G}{\mathrm{std}} \Big[ r^{(k)}_{t'_{j}} \Big]}
\end{equation}

In practice, early experiments showed that using a regular frame indexes pattern along the input speech content performed better. Thus we introduce $n_w \in \mathbb{N}^*$ and use the end timestamp of every $n_w$ words in the input to set $(t'_j)_{j \le s}$.

Then, we introduce the advantage of an output at step $t$ as the sum of normalized rewards from the following steps:
\begin{equation}
\label{eq:advantage}
R^{(i)}_t = \sum\limits_{t'_j > t} \bar{r}^{(i)}_{t'_{j}}
\end{equation}

We compute the per-codebook objectives $L^{(i)}_{q}$ using the standard clipping function between $1 - \epsilon$ and $1 + \epsilon$ as:
\begin{equation}
\label{eq:intermediate_objective}
L^{(i)}_{q} = \sum\limits_{t=1}^{T} \min \Big( p^{(i)}_{q,t} R^{(i)}_t, \mathrm{clip}_{\epsilon} \big( p^{(i)}_{q,t} \big) R^{(i)}_t \Big)
\end{equation}

In the end, we seek to maximize the following objective with fixed weights $c_q$ for each depth $q \le Q$:
\begin{equation}
\label{eq:grpo_objective}
\E_{\substack{
X \sim \mathcal{D} \\
Y_i \sim \pi_{\theta_{\mathrm{old}}}(X)
}}
\left[
\frac{1}{G} \sum_{\substack{q \le Q \\ i \le G}} c_q \, L^{(i)}_{q}
\right]
\end{equation}

where $\mathcal{D}$ denotes our input speech distribution and $\pi_{\theta_{\mathrm{old}}}$ is a fixed version of the translation model that is replaced by $\pi_{\theta}$ every fixed number of updates $\tau$.

\section{Experiments}
% \label{sec:experiments}

%%% SECTION %%%
\subsection{Architectural hyper-parameters}
\label{sec:arch_hyper_params}
The backbone \emph{Temporal} Transformer of \ours{} has a latent dimension of 2048 (8192 for the SiLU gating), 28 layers, 16 heads and local attention over 3000 tokens, \emph{i.e.}, 2B parameters and a 4min context. The \emph{Depth} Transformer has a latent dimension of 1024 (4096 for the gating), 6 layers per codebook and 16 heads. It models $Q=16$ audio codebooks for the output stream and the same for the input stream but only at training. We reduce the size of the model before RL by distillation into a smaller one using weight sharing among the codebooks of the \emph{Depth} Transformer. Our final model architecture contains 3B parameters.

\begin{table*}[t]
    \caption{Objective comparison of \ours{} with Seamless~\citep{seamless} and Hibiki~\citep{hibiki} on short-form (Europarl-ST) and long-form (Audio-NTREX-4L) test data introduced in Section~\ref{sec:eval_datasets}.}
    \label{tab:objective_results}
    % \vskip 0.15in
    \begin{center}
    \begin{sc}
    \resizebox{0.975 \linewidth}{!}{%
    \begin{tabular}{lcccccccccccc}
    \toprule
     & \multicolumn{6}{c}{Short-form} & \multicolumn{6}{c}{Long-form} \\
    \cmidrule(lr){2-7}\cmidrule(lr){8-13}
          &                   & ASR               & ASR & Speaker           & End & & & ASR & ASR & Speaker & End & \\
    & BLEU ($\uparrow$) & BLEU ($\uparrow$) & COMET ($\uparrow$) & Sim. ($\uparrow$) & Offset ($\downarrow$) & LAAL ($\downarrow$) & BLEU ($\uparrow$) & BLEU ($\uparrow$) & COMET ($\uparrow$) & Sim. ($\uparrow$) & Offset ($\downarrow$)  & LAAL ($\downarrow$) \\
    \midrule
    \textbf{Seamless} \\
    French
        & 33.8 & 32.8 & 76.6 & 19.1 & 2.4 & \textbf{2.8}
        & 27.8 & 23.9 & 33.7 & 44.4 & 3.2 & 6.2 \\
    Spanish
        & \textbf{34.4} & 33.6 & 79.1 & 21.9 & 2.6 & \textbf{2.7}
        & 29.9 & 25.2 & 36.1 & 42.6 & 2.8 & 6.5 \\
    Portuguese
        & \textbf{34.1} & \textbf{33.6} & 78.9 & 23.9 & 2.8 & 3.1
        & 29.0 & 25.6 & 35.0 & 35.7 & 3.2 & 6.6 \\
    German
        & 27.8 & 27.3 & \textbf{82.3} & 20.6 & 2.4 & 3.0
        & 27.8 & 24.0 & 40.6 & 47.8 & 2.5 & 7.3 \\
    \midrule
    \textbf{Hibiki} \\
    French
        & 32.4 & 31.8 & \textbf{81.5} & 35.7 & 2.5 & 3.5
        & 29.5 & 26.4 & 42.0 & 52.8 & 2.6 & 6.8 \\  %
    \midrule
    \textbf{\ours{}} \\  % DISTIL 77f82164@110
    French
        & \textbf{35.0} & \textbf{34.6} & 80.3 & \textbf{49.5} & \textbf{2.1} & \textbf{2.8}
        & \textbf{30.6} & \textbf{28.7} & \textbf{43.7} & \textbf{61.3} & \textbf{2.3} & \textbf{6.1} \\
    Spanish
        & 33.8 & \textbf{33.9} & \textbf{80.3} & \textbf{57.0} & \textbf{2.3} & 3.1
        & \textbf{32.3} & \textbf{31.5} & \textbf{42.3} & \textbf{64.6} & \textbf{2.6} & \textbf{5.6} \\
    Portuguese
        & 33.6 & \textbf{33.6} & \textbf{78.9} & \textbf{51.4} & \textbf{2.4} & \textbf{3.0}
        & \textbf{33.2} & \textbf{31.3} & \textbf{42.6} & \textbf{62.1} & \textbf{2.3} & \textbf{6.3} \\
    German
        & \textbf{28.7} & \textbf{28.6} & 82.0 & \textbf{51.5} & \textbf{1.9} & \textbf{2.8}
        & \textbf{29.1} & \textbf{28.3} & \textbf{42.3} & \textbf{66.0} & \textbf{2.0} & \textbf{5.9} \\
    \bottomrule
    \end{tabular}}
    \end{sc}
    \end{center}
    % \vskip -0.1in
\end{table*}

%%% SECTION %%%
\subsection{Training protocol}
\label{sec:training_protocol}
We train a multilingual-to-English speech translation system through the following steps, each with a cosine learning rate schedule and AdamW~\citep{adamw}, with weight decay of 0.1, and momentum parameters (0.9, 0.95).

\subsubsection{Text backbone initialization}
\label{sec:text_pretraining}
We initialize the \emph{Temporal} Transformer with Helium-1\footnote{\href{https://huggingface.co/kyutai/helium-1-2b}{huggingface.co/kyutai/helium-1-2b}} \citep{helium} weights, an open-source base text LLM with 2B parameters trained using filtered Common Crawl\footnote{\href{https://commoncrawl.org}{commoncrawl.org}} data.

\subsubsection{Audio pretraining}
\label{sec:audio_pretraining}
Starting from the pretrained text backbone, weights of a \emph{Depth} Transformer modeling 8 audio codebooks are added to the architecture as well as audio tokens projection layers. We perform an audio pretraining with single stream audio as done by \citet{hibiki} but on multilingual speech. Our data mixture comprises approximately 12\% of audio in each input language, 50\% of English and less than 2\% of Italian. We train for 1M steps with a batch size of 144 and a learning rate of $2\cdot10^{-4}$. After this pretraining stage, we add new randomly initialized weights to the \emph{Depth} Transformer corresponding to 8 additional audio codebooks so it can indeed model a total of $Q=16$ audio codebooks in the next training stages. We then duplicate all the weights of the \emph{Depth} Transformer to allow for future multistream training i.e. modeling $Q=16$ audio codebooks for the input stream as well as the output stream. Appendix Figure~\ref{fig:ce_pretraining} shows the cross-entropy of output text and audio tokens in pretraining.

\subsubsection{Coarse speech translation training}
\label{sec:st_training}

We construct a large-scale multilingual-to-English speech translation dataset comprising $40,000$ hours for each source language (French, Spanish, Portuguese, and German). Starting from a massive collection of multilingual audio, we extract 4 million single-speaker utterances, whose durations are between 30 and 75 seconds, and transcribe them using Whisper large-v3~\citep{whisper}. Transcripts are partitioned into sentences via Spacy's \texttt{core\_news\_sm} and individually translated using MADLAD-3B~\citep{madlad}, after which we synthesize the target speech using the TTS system described in Section~\ref{sec:nat_pauses_tts} with 10-second speaker conditioning. To ensure coarse translation alignments, we apply the silence insertion technique from Section~\ref{sec:sent_level_align} using $\delta=0.5$ and $\mu=2$. Scaling our training budget following \citet{hibiki}, we perform $500,000$ gradient steps with a batch size of 96 and a learning rate of $3\cdot10^{-5}$, computing the loss on both source and target streams with source noise augmentation. Finally, sequence termination is explicitly modeled by inserting a special input EOS token immediately following the source utterance and a separate EOS token in the text stream to indicate the end of generation. Appendix Figure~\ref{fig:ce_sft} shows cross-entropy curves for output text and audio tokens during training, while Appendix Table~\ref{tab:multi_vs_monolingual} compares multilingual and monolingual model performance.

\subsubsection{Speech translation fine-tuning}
\label{sec:st_finetuning}
We use the synthetic data generation method with natural pauses introduced in Section~\ref{sec:nat_pauses_tts} to build a synthetic multilingual speech translation dataset of less than 200h in total. We fine-tune for 1K steps with a batch size of 16, a learning rate of $1\cdot10^{-6}$ and other configurations being similar to the previous phase described in \ref{sec:st_training}. We then distill the model into a light copy of itself with codebooks weight sharing using the same dataset and 20K gradient updates.

\subsubsection{Reinforcement learning}
\label{sec:st_rl}
Starting from the light fine-tuned translation model, we use data from the speech translation training introduced in Section~\ref{sec:st_training} and run our reinforcement learning process as described in Section~\ref{sec:rl_method}. We train with a batch size of 32, a group size of 4, learning rate of $2\cdot10^{-7}$ and perform 2000 updates with $\tau=20$. Sequences of length $T=1500$ frames are generated using a temperature of 0.8 and top-k of 250 for both text and audio streams. Process rewards are computed every $n_w=8$ input words and we set $\alpha=0.4$ and $\epsilon=0.2$. We use $c_0=100$ and $c_q=1$ for $q \ge 1$ to balance loss between text and audio streams. The model is evaluated every $10\cdot\tau$ updates on a valid set and we define Hibiki-Zero as the checkpoint with the best quality/latency trade-off according to objective evaluation metrics. Appendix Table~\ref{tab:base_and_finetuned_results} compares our base and fine-tuned models to \ours{}.

%%% SECTION %%%
\subsection{Evaluation datasets}
\label{sec:eval_datasets}

\paragraph{Long-form data.}
We build Audio-NTREX-4L, a multilingual long-form ST dataset using text translations from the NTREX~\cite{ntrex} corpus. We select 300 examples for each source language and synthesize them using the following high-quality TTS from the industry: ElevenLabs\footnote{\href{https://elevenlabs.io/text-to-speech}{elevenlabs.io/text-to-speech} (``eleven-multilingual-v2 TTS'')}, Cartesia\footnote{\href{https://cartesia.ai/sonic}{cartesia.ai/sonic} (``sonic-v2 TTS'')} and Gradium\footnote{\href{https://gradium.ai/\#models}{gradium.ai/\#models} (``default TTS'')}. We condition generations using voices from the multilingual dataset CML-TTS~\citep{cml_tts}. Audio-NTREX-4L contains around 15h of speech per TTS with an average duration of 45 seconds per sample and is split in balanced valid and test sets.
% \footnote{\href{https://www.openslr.org/146/}{openslr.org/146/}}

\paragraph{Short-form data.}
We filter data from Europarl-ST~\citep{europarl_st} and retain samples with realistic transcripts and duration between 2 and 20 seconds. We build valid and test sets, each with 1024 samples per source language for a total of 10h hours per set.
% \footnote{\href{https://mllp.upv.es/europarl-st/}{mllp.upv.es/europarl-st/}}

%%% SECTION %%%
\subsection{Evaluation metrics}
\label{sec:eval_metrics}

\looseness=-1
\paragraph{Translation quality.} We evaluate translation quality by transcribing generated speech using Whisper medium~\cite{whisper} and computing BLEU~\cite{sacrebleu} and COMET~\cite{comet} scores with respect to a reference translation, referred to as ASR-BLEU and ASR-COMET. To reduce the impact of ASR errors, hypothesis and reference texts are normalized\footnote{\href{https://github.com/openai/whisper/blob/main/whisper/normalizers/english.py}{github.com/openai/whisper/blob/main/whisper/normalizers}} before computing BLEU scores. Since Seamless and \ours{} perform speech-to-text translation in parallel, we also compute BLEU and COMET scores using their text outputs. We use the XCOMET-XL model.\footnote{\href{https://github.com/Unbabel/COMET}{github.com/Unbabel/COMET}}.

\looseness=-1
\paragraph{Translation Latency.} We rely on two common latency metrics known as End Offset and LAAL (Length-Adaptive Average Lagging). End Offset is defined as the time difference (in seconds) between the end of the last generated word and the end of the last word from the source. We compute LAAL following the method described by \citet{laal} which defines it as an approximation of the average time (in seconds) between a source word and its translation. We use word-level emission timestamps $(d_i)_{1 \dots n_{\mathrm{gen}}}$ produced by Whisper for $n_{\mathrm{gen}}$ words in the generated speech. We define $\gamma = \frac{\Delta_{\mathrm{source}}}{\max(n_{\mathrm{gen}}, n_{\mathrm{ref}})}$ where $\Delta_{\mathrm{source}}$ is the duration of the source speech and $n_{\mathrm{ref}}$ the number of words in the reference translation. The LAAL score is then computed as $\frac{1}{n_{\mathrm{max}}}\sum_{i=1}^{n_{\mathrm{max}}} d_i - (i-1)\gamma$ where $n_{\mathrm{max}} = \mathrm{min}\{i|d_i \geq \Delta_{\mathrm{source}}\}$.

\begin{table}[t]
    \caption{\textbf{Human evaluation.} Raters report Mean Opinion Scores (MOS) on a scale ranging from 0 to 100 for each audio sample.}
    \label{tab:human_eval}
    \begin{center}
    \begin{scriptsize}
    \begin{sc}
    \begin{tabular}{@{}ll@{\hspace{5pt}}c@{\hspace{5pt}}c@{\hspace{5pt}}c@{}}
    \toprule
    Input    & \multirow{2}{*}{Model} & Audio   & Speaker    & Speech \\
    language &                        & Quality & Similarity & Naturalness \\
    \midrule
    \multirow{3}{*}{French} & Seamless & 11.4 $\pm$ 3.1 & 21.1 $\pm$ 4.9 & 21.2 $\pm$ 3.8 \\
                            & Hibiki & 62.9 $\pm$ 4.8 & 44.7 $\pm$ 5.1 & 57.0 $\pm$ 4.2 \\
                            & \ours{} & \textbf{64.5} $\pm$ 4.2 & \textbf{70.0} $\pm$ 5.1 & \textbf{67.2} $\pm$ 4.1 \\
    \midrule
    \multirow{2}{*}{Spanish} & Seamless & 10.7 $\pm$ 2.6 & 21.2 $\pm$ 4.5 & 26.5 $\pm$ 4.4 \\
                             & \ours{} & \textbf{66.8} $\pm$ 3.9 & \textbf{69.0} $\pm$ 3.9 & \textbf{66.2} $\pm$ 4.9 \\
    \midrule
    \multirow{2}{*}{Portuguese} & Seamless & 11.8 $\pm$ 3.1 & 32.5 $\pm$ 6.0 & 22.8 $\pm$ 3.9 \\
                                & \ours{} & \textbf{62.0} $\pm$ 4.1 & \textbf{60.7} $\pm$ 4.2 & \textbf{75.6} $\pm$ 3.4 \\
    \midrule
    \multirow{2}{*}{German} & Seamless & 15.6 $\pm$ 2.7 & 25.2 $\pm$ 4.9 & 26.4 $\pm$ 4.8 \\
                            & \ours{} & \textbf{73.5} $\pm$ 3.4 & \textbf{65.3} $\pm$ 4.3 & \textbf{69.9} $\pm$ 3.9 \\
    \bottomrule
    \end{tabular}
    \end{sc}
    \end{scriptsize}
    \end{center}
\end{table}

\begin{table}[t]
    \caption{Objective results of model adaptation to input Italian speech with 850 hours of finetuning data on short-form evaluation.}
    \label{tab:italian_results}
    \begin{center}
    \setlength{\tabcolsep}{3pt}
    \begin{scriptsize}
    \begin{sc}
    \begin{tabular}{lccccc}
    \toprule
    & BLEU                  & ASR               & Speaker           & End                   & LAAL \\
    &          ($\uparrow$) & BLEU ($\uparrow$) & Sim. ($\uparrow$) & Offset ($\downarrow$) & ($\downarrow$) \\
    \midrule
    \textbf{Seamless} & \textbf{32.5} & \textbf{32.0} & 22.2 & \textbf{3.0} & \textbf{3.5} \\
    \midrule
    \textbf{Ours} \\
    Base & 14.3 & 14.3 & 50.6 & 3.9 & 4.3 \\  % 5c94ad8f@500
    Finetuned & 31.4 & 31.0 & \textbf{55.2} & 3.7 & 4.5 \\  % 6f8df93b@200
    Finetuned + RL & 32.1 & 31.9 & 54.2 & \textbf{3.0} & \textbf{3.5} \\  % a475c272@100
    \bottomrule
    \end{tabular}
    \end{sc}
    \end{scriptsize}
    \end{center}
\end{table}

\looseness=-1
\paragraph{Cross-lingual speaker similarity.} For objective voice transfer evaluation, we use a standard model for speaker verification\footnote{\href{https://github.com/microsoft/UniSpeech/tree/main/downstreams/speaker\_verification\#pre-trained-models}{github.com/microsoft/UniSpeech} (``WavLM Large'')} based on WavLM~\cite{wavlm} and report the cosine similarity between the embeddings of the source and the generated speech.

\looseness=-1
\paragraph{Audio quality and naturalness.} We rely on human raters to evaluate audio quality, speech naturalness and additional cross-lingual speaker similarity of generated audios. We conduct evaluations per input language using 50 samples and 20 raters for each model with 5 comparisons per rater.
% The detailed protocol is given in Appendix.

%%% SECTION %%%
\subsection{Inference configuration}
\label{inference_config}
We encode audio with the streaming codec and feed the tokens to \ours{} while decoding the output tokens to obtain a streaming translation. At the end of the input, we force EOS tokens to our model input audio streams, and keep sampling until it produces its own text stream EOS. We use temperature of 0.8 and top-k of 250 for all tokens.

%%% SECTION %%%
\subsection{Results}

\paragraph{Objective evaluations.}
\label{sec:objective_st_results}
Table~\ref{tab:objective_results} compares \ours{} against the best available baselines for simultaneous S2ST namely Seamless~\citep{seamless} and Hibiki~\citep{hibiki} with the latter only supporting French as input. Our model outperforms both baselines on long-form speech translation with more than 5pts of ASR BLEU, 20pts of speaker similarity and lower latency compared to Seamless. In the short-form setting, our approach outperforms Hibiki by 3pts of ASR BLEU while being faster and is on par with Seamless on the quality/latency trade-off but surpasses it on speaker similarity by more than 30pts.

\paragraph{Audio fidelity and speech expressivity.}
\label{sec:human_eval_results}
Human evaluations reported in Table~\ref{tab:human_eval} confirm the clear advantage of \ours{} compared to Seamless on speaker identity transfer but also show that it produces higher quality audio with better speech naturalness. Compared to Hibiki on a French-to-English task, our model reaches equivalent audio quality while being more natural with a better speaker similarity.

\paragraph{New language adaptation.}
\label{sec:new_lang_results}
Following our method from Section~\ref{sec:st_training}, we build a small coarse-aligned Italian-to-English ST dataset containing less than 1000 hours in each language. Starting from the base translation model obtained after the training stage described in Section~\ref{sec:st_training}, we fine-tune and apply our RL method for the Italian-to-English translation task only. Results are presented in Table~\ref{tab:italian_results} and show that we attain the same translation quality/latency trade-off as Seamless with better speaker similarity on an extension to Italian of our short-form evaluation data. As shown in Appendix Table~\ref{tab:italian_results_longform}, our model adapted to Italian also retains most of its capabilities on the original languages.

\begin{figure}[t]
    \centering
    \includegraphics[width=\columnwidth]{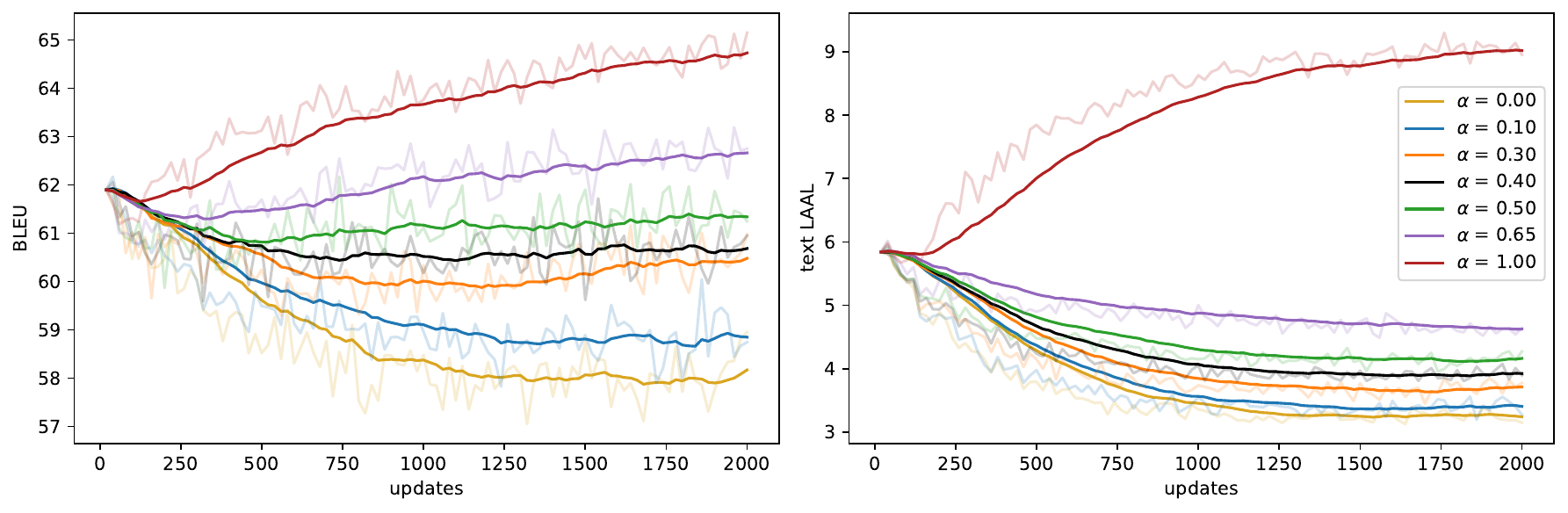}
    \caption{\textbf{Influence of hyperparameter $\alpha$ during RL.} We plot the BLEU score and text LAAL over training for various $\alpha$ (see Eq.~\eqref{eq:process_reward}), starting from the same supervised model using $n_w=8$.}
    \label{fig:ablation_alpha}
\end{figure}

%%% SECTION %%%
\subsection{Ablations}
\label{sec:ablations}

\begin{figure}[t]
    \centering
    \includegraphics[width=\columnwidth]{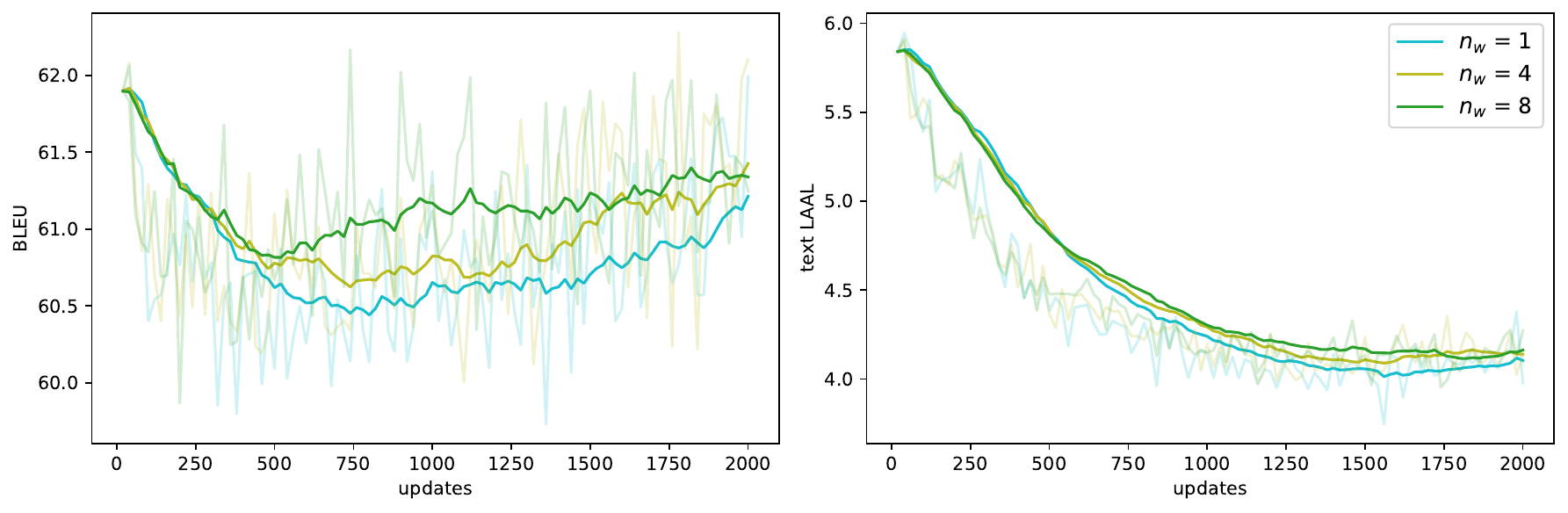}
    \caption{\textbf{Influence of hyperparameter $n_w$ during RL.} We plot the BLEU score and text LAAL over training for various $n_w$ (see Sec.~\ref{sec:optim_objective}) starting from the same supervised model using $\alpha=0.5$.}
    \label{fig:ablation_fw}
\end{figure}

\begin{figure}[t]
    \centering
    \includegraphics[width=\columnwidth]{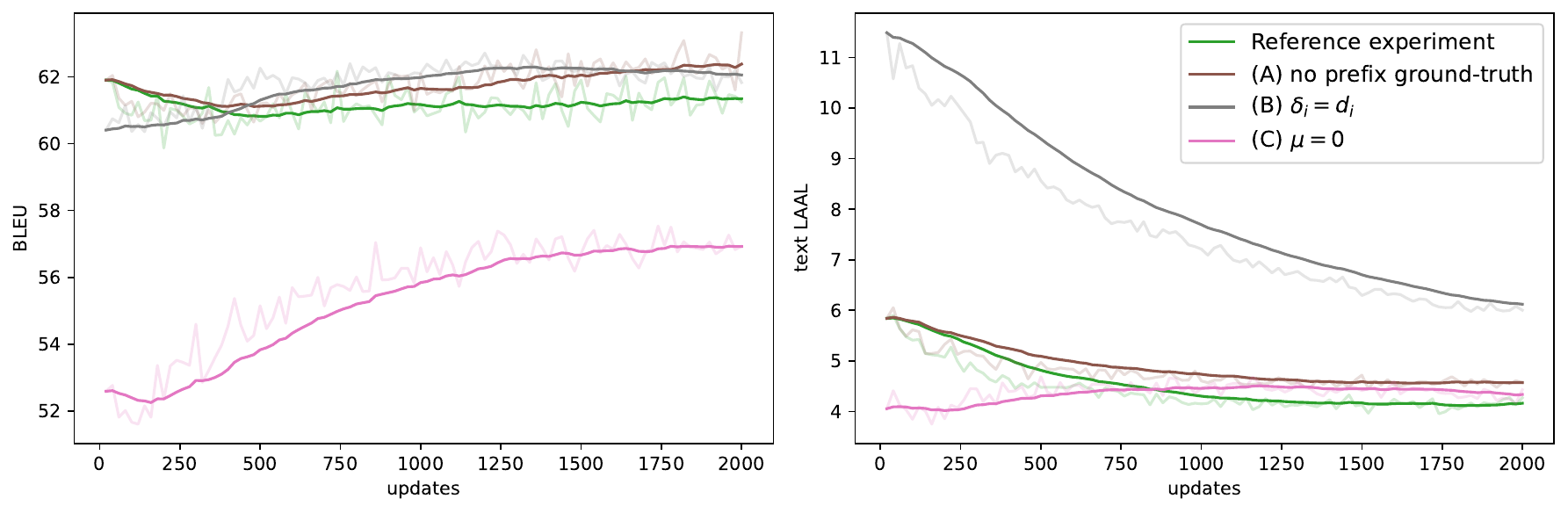}
    \caption{\textbf{Alternative configurations.} We use $\alpha=0.5$ and $n_w=8$ for all experiments. Experiment (A) uses the full translation of the input speech as reference to compute process rewards instead of sentence-level prefixes as in Equation~\ref{eq:process_reward}. Experiment (B) performs RL on a supervised model trained with full sentence-delay i.e. $\delta_i=d_i$ for each input sentence of index $i$. Experiment (C) performs RL on a supervised model trained with coarse alignments using silences between sentences only i.e. $\mu=0$.}
    \label{fig:ablation_misc_setup}
\end{figure}

We present ablation results in figures~\ref{fig:ablation_alpha}, \ref{fig:ablation_fw} and \ref{fig:ablation_misc_setup} using exponential moving average smoothing for readability. Performance during RL is represented using BLEU and text LAAL metrics. They are computed every $\tau$ updates on a validation set using the output text stream of the model. As observed by~\citet{hibiki}, we also notice very high BLEU scores ($\approx 60$) compared to evaluation scores ($\approx 30$). Indeed, our train and valid sets were obtained with the same data generation process described in Section~\ref{sec:st_training} thus following the same translation style as MADLAD-3B~\citep{madlad} that our ST models learn to replicate.

\begin{figure}[t]
    \centering
    \includegraphics[width=\columnwidth]{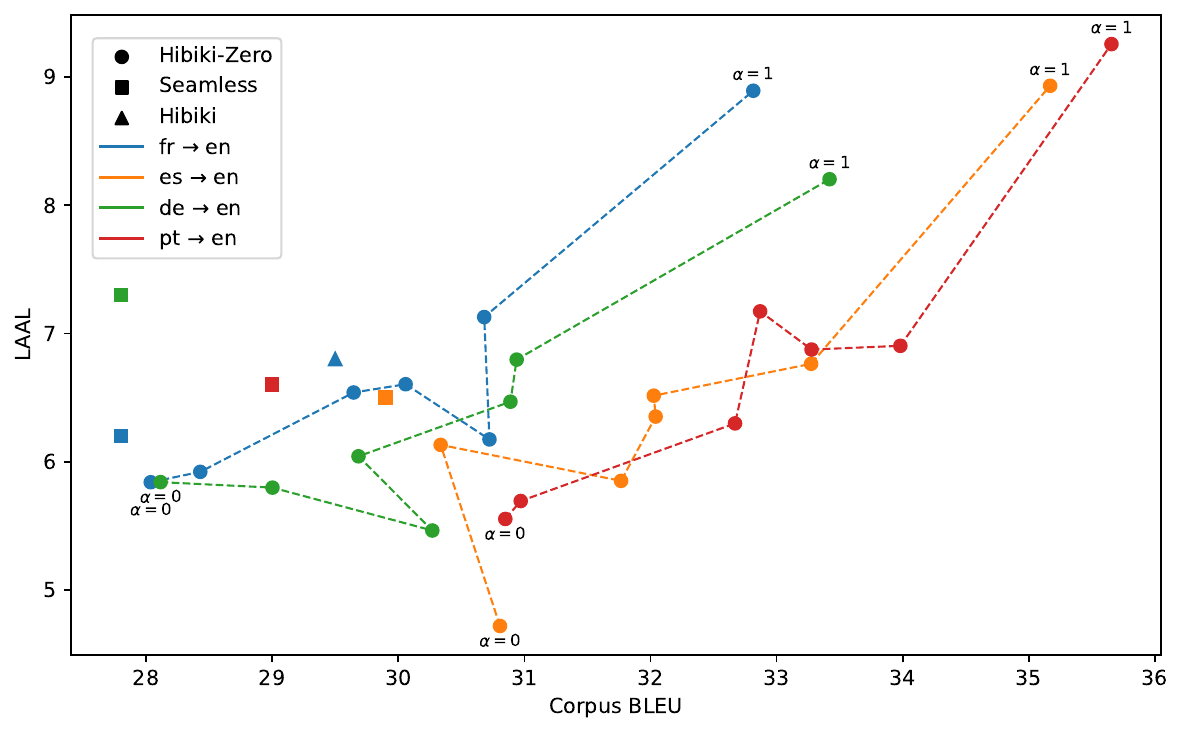}
    \caption{\textbf{Comparison of translation quality/latency trade-off.} We evaluate our approach by training with various values of $\alpha$ in $\{0.0, 0.1, 0.3, 0.4, 0.5, 0.65, 1.0\}$ and compare against baselines on the long-form evaluation benchmark Audio-NTREX-4L.}
    \label{fig:ql_tradeoff}
\end{figure}

\paragraph{Ablation: Quality/Latency control during RL.}
\label{sec:ablation_alpha}
We benchmark the effect of parameter $\alpha$ introduced in Section~\ref{sec:process_rewards} which balances total and intermediate BLEU scores in the definition of process rewards. As illustrated in Figure~\ref{fig:ablation_alpha}, performing RL with high values of $\alpha$ leads to a higher translation latency but better overall translation quality as expected. On the contrary, lower values of $\alpha$ reduce latency further at the cost of a limited quality decrease. In Figure~\ref{fig:ql_tradeoff}, we present quality/latency trade-off curves across languages pairs. Example translations for different values of $\alpha$ are provided in Appendix Figures~\ref{fig:fr_outputs}, \ref{fig:es_outputs}, \ref{fig:pt_outputs}, and \ref{fig:de_outputs}.

\paragraph{Ablation: Process rewards computation frequency.}
\label{sec:ablation_fw}
Using $\alpha=0.5$, we study the effect of parameter $n_w$ introduced in Section~\ref{sec:optim_objective} which controls how often process rewards are computed along a generated sequence. As shown in Figure~\ref{fig:ablation_fw}, decreasing this parameter below $8$ does not significantly impact the final quality/latency trade-off.

\begin{figure}[t]
    \centering
    \includegraphics[width=\columnwidth]{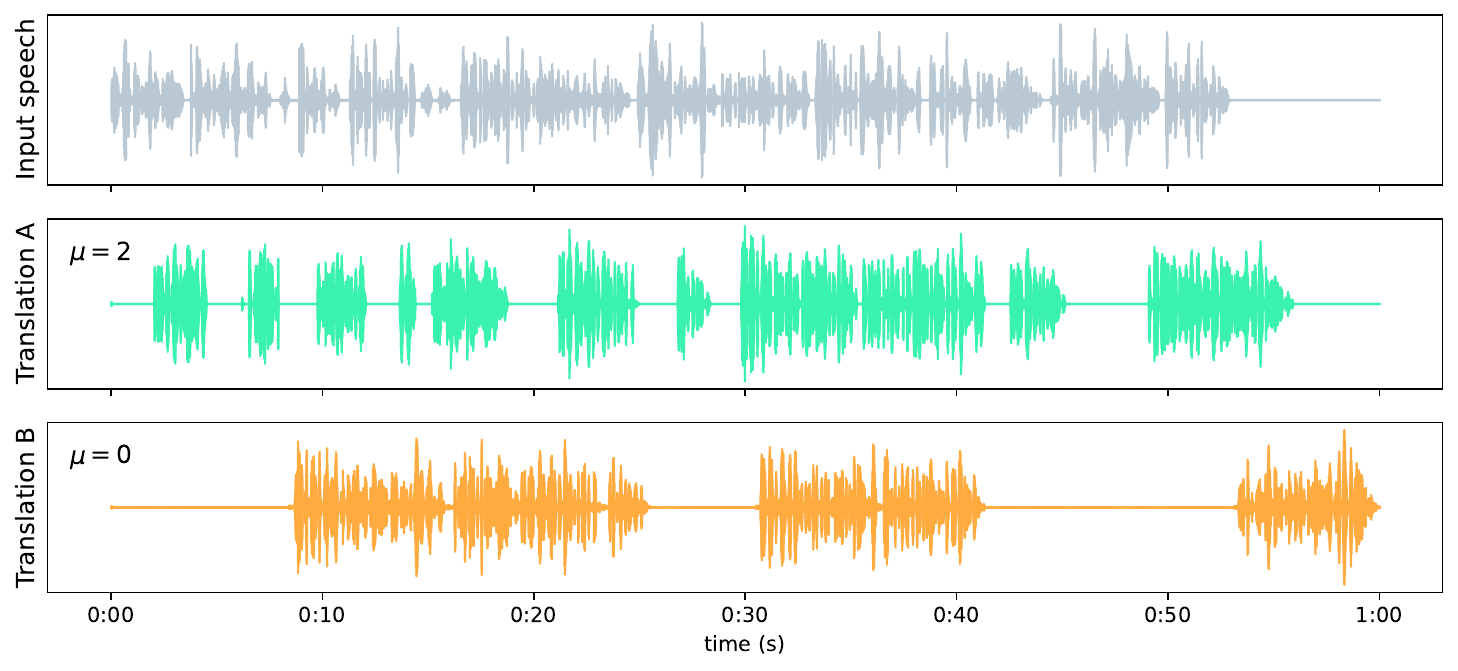}
    \caption{\textbf{Illustration of coarse translation alignment patterns} Waveform A is generated by a model trained on coarse alignments with random silences. Waveform B is generated by a model trained on coarse alignments with silences between sentences only.}
    \label{fig:baseline_silences_pattern}
\end{figure}

\paragraph{Ablation: Alternative configurations.}
\label{sec:ablation_misc}

In Figure~\ref{fig:ablation_misc_setup}, we compare alternative configurations that could be used for model development instead of our main setup referred to as \textit{Reference experiment}. We keep $\alpha=0.5$ and $n_w=8$ fixed. \\

Experiment (A) performs RL using the full translation of the reference input text instead of sentence-level prefixes to compute intermediate BLEU scores. This amounts to modify Equation~\ref{eq:process_reward} so it becomes $r^{(i)}_{\mathrm{noprefix},t} = (1 - \alpha) \mathrm{BLEU} ( \hat{y}^{(i)}_{t}, y_{T} ) + \alpha \mathrm{BLEU} ( \hat{y}^{(i)}_{T}, y_{T} )$. We observe better quality performance but at the cost of latency. According to us, this comes from intermediate BLEU scores being much noisier as translated references are too optimistic, making it harder to discriminate between sequences to optimize latency during RL. \\

Experiment (B) performs RL starting from a base model trained with full sentence delays between input and output speech meaning that $\delta_i=d_i$ for each sentence index $i$ using notations from Section~\ref{sec:sent_level_align}. Therefore, latency is much higher when starting RL and is reduced to around 6 seconds which remains far worse than the reference experiment. We also observe this behavior in preliminary experiments where RL was unable to teach the base model to start translation of an input sentence before it ends as the base model was never trained in that manner during supervised training. This justifies the use of $\delta_i<d_i$ when building coarse alignments so RL can benefit from exploration. \\

\vspace{-1em}
Experiment (C) performs RL starting from a base model trained with coarse alignments using sentence-level silences only ($\mu=0$). We observe a degradation both in terms of quality and latency compared to the reference experiment. The loss of quality is expected when decreasing $\mu$ as we don't delay as much the output with respect to the input. The cause of higher latency is illustrated in Figure~\ref{fig:baseline_silences_pattern} where waveform B ($\mu=0$) is a speech translation where silences are located between sentences only. This results in a higher average latency than waveform A ($\mu>0$) which presents a better distribution of speech along time.

%%% SECTION %%%
\subsection{Limitations}
\label{sec:limitations}

\looseness=-1
This work proposes an efficient method to perform multilingual speech translation and shows promising results on new input language adaptation.
However, while our model exhibits state-of-the-art speaker identity preservation, there is no way to control the intensity of the accent from the input language in the generated speech. Such control could be added by providing accent-annotated samples during supervised training and using conditioning at inference. Similarly, while the reinforcement learning stage enables steering the model toward a particular quality/latency trade-off, this trade-off cannot be adjusted at inference time.
\section{Conclusion}
\label{sec:conclusion}

\looseness=-1
We present \ours{}, a multilingual model for simultaneous and expressive speech and text translation without requiring word-level alignment of translation data for training. Our method leverages coarse sentence-level alignments to train a base model that is further refined through Reinforcement Learning using process rewards based on BLEU score only. \ours{} outperforms the state-of-the-art across multiple languages with better quality/latency trade-offs, speaker identity transfer and speech naturalness. Moreover, we demonstrate new language adaptation with our method using less than 1000 hours of speech data. We release \ours{} weights as well as our multilingual long-form evaluation dataset to benefit the research community.

\section*{Impact Statement}
\looseness=-1
Our work aims at making simultaneous speech translation more accessible in multiple languages, by improving its quality but also by facilitating the research and development effort to build it. This technology has potential to improve human communication. While speech translation may impact interpreter employment in settings such as international meetings or live interviews, it is likely to have a net positive effect in contexts where interpreters are not typically available. We also acknowledge risks related to voice transfer but indicate that speech translation with voice transfer presents a lower spoofing risk than standard text-to-speech systems, as the output speech content is not directly controllable and the model’s freedom to reformulate inputs is limited.

\bibliography{references}
\bibliographystyle{icml2026}

\newpage
\appendix
\onecolumn
\section*{Appendix}
\label{sec:appendix}

\begin{table*}[h]
    \caption{\textbf{Objective evaluations of multilingual and monolingual base supervised models.} As the multilingual base model is trained on four times more data than each monolingual model, it has seen the same amount of each language after 400K updates as any monolingual model after 100K updates. For comparison, we provide evaluations after 100K and 400K updates for the multilingual model.}
    \label{tab:multi_vs_monolingual}
    \begin{center}
    \begin{sc}
    \resizebox{0.975\linewidth}{!}{
    \begin{tabular}{lcccccccccccc}
    \toprule
     & \multicolumn{5}{c}{Short-form} & \multicolumn{5}{c}{Long-form} \\
    \cmidrule(lr){2-6}\cmidrule(lr){7-11}
    
    \multirow{2}{*}{Model} & & ASR & Speaker & End & & & ASR & Speaker & End & \\
                           & BLEU ($\uparrow$) & BLEU ($\uparrow$) & Sim. ($\uparrow$) & Offset ($\downarrow$) & LAAL ($\downarrow$) & BLEU ($\uparrow$) & BLEU ($\uparrow$) & Sim. ($\uparrow$) & Offset ($\downarrow$) & LAAL ($\downarrow$) \\
    
    \midrule
    \textbf{French} \\
    Base@100K
        & 31.8 & 31.3 & 52.1 & 3.1 & \textbf{3.6}
        & 28.7 & 27.7 & \textbf{67.6} & 3.4 & \textbf{6.0} \\
    Base@400K
        & \textbf{34.5} & \textbf{34.1} & \textbf{53.1} & \textbf{3.0} & 3.7
        & \textbf{31.1} & \textbf{28.8} & 67.4 & 3.4 & 6.6 \\
    Base-FR@100K
        & 34.2 & 33.4 & 52.5 & \textbf{3.0} & \textbf{3.6}
        & 29.8 & 28.6 & 67.3 & \textbf{3.2} & 6.1 \\
    Base-ES@100K
        & 5.3 & 5.1 & 39.6 & 4.4 & 5.1
        & 5.9 & 5.5 & 60.1 & 7.7 & 11.2 \\
    Base-PT@100K
        & 6.7 & 6.6 & 41.2 & 4.6 & 5.0
        & 7.9 & 8.0 & 62.0 & 5.8 & 8.5 \\
    Base-DE@100K
        & 1.6 & 1.6 & 41.6 & 4.3 & 5.6
        & 2.0 & 1.7 & 64.1 & 8.3 & 10.8 \\
    
    \midrule
    \textbf{Spanish} \\
    Base@100K
        & 31.5 & 31.4 & 59.2 & \textbf{3.4} & \textbf{4.1}
        & 31.1 & 30.9 & 69.5 & 3.9 & \textbf{6.6} \\
    Base@400K
        & \textbf{33.8} & \textbf{33.6} & \textbf{60.3} & \textbf{3.4} & \textbf{4.1}
        & 33.3 & 32.6 & 69.4 & \textbf{3.8} & \textbf{6.6} \\
    Base-FR@100K
        & 8.9 & 8.8 & 48.2 & 3.8 & \textbf{4.1}
        & 11.4 & 11.4 & 63.6 & 4.6 & 7.3 \\
    Base-ES@100K
        & 33.2 & 33.0 & 59.9 & 3.5 & 4.2
        & \textbf{33.5} & \textbf{32.7} & \textbf{69.8} & 4.1 & 6.6 \\
    Base-PT@100K
        & 22.3 & 22.2 & 50.2 & 3.5 & \textbf{4.1}
        & 24.5 & 24.4 & 63.8 & 4.2 & 7.0 \\
    Base-DE@100K
        & 1.3 & 1.1 & 38.9 & 4.2 & 5.1
        & 2.1 & 1.7 & 57.1 & 9.6 & 12.3 \\
    
    \midrule
    \textbf{Portuguese} \\
    Base@100K
        & 31.7 & 31.4 & 53.5 & 3.7 & 4.2
        & 31.7 & 30.7 & 66.9 & 3.3 & 6.6 \\
    Base@400K
        & \textbf{33.9} & \textbf{33.7} & \textbf{54.5} & \textbf{3.6} & \textbf{4.1}
        & \textbf{33.5} & \textbf{32.5} & 67.4 & \textbf{3.1} & \textbf{6.5} \\
    Base-FR@100K
        & 2.5 & 2.4 & 43.2 & 4.2 & 4.7
        & 9.0 & 8.6 & 53.6 & 4.9 & 7.8 \\
    Base-ES@100K
        & 12.8 & 12.8 & 47.3 & 4.2 & 4.9
        & 23.9 & 23.0 & 56.9 & 3.8 & 8.0 \\
    Base-PT@100K
        & 32.5 & 32.2 & 54.4 & 3.9 & 4.4
        & 32.2 & 30.8 & \textbf{67.8} & 3.4 & 7.0 \\
    Base-DE@100K
        & 0.6 & 0.7 & 42.2 & 3.9 & 5.1
        & 1.3 & 0.9 & 52.0 & 10.8 & 11.7 \\
    
    \midrule
    \textbf{German} \\
    Base@100K
        & 25.9 & 25.7 & 53.6 & \textbf{2.7} & \textbf{3.5}
        & 28.3 & 28.0 & 70.6 & 3.4 & 6.4 \\
    Base@400K
        & 28.3 & 28.0 & 54.6 & \textbf{2.7} & 3.6
        & \textbf{31.1} & \textbf{30.5} & 70.5 & \textbf{3.3} & \textbf{5.8} \\
    Base-FR@100K
        & 0.7 & 0.7 & 35.6 & 4.4 & 4.6
        & 1.4 & 1.2 & 54.5 & 12.1 & 10.0 \\
    Base-ES@100K
        & 0.9 & 0.8 & 35.4 & 4.6 & 5.2
        & 1.8 & 1.1 & 50.9 & 14.5 & 16.0 \\
    Base-PT@100K
        & 0.7 & 0.7 & 33.3 & 4.1 & 4.6
        & 1.6 & 1.4 & 50.2 & 8.1 & 9.6 \\
    Base-DE@100K
        & \textbf{28.6} & \textbf{28.3} & \textbf{55.6} & 2.9 & 3.8
        & 30.8 & 29.9 & \textbf{71.4} & 3.4 & 6.2 \\
        
    \bottomrule
    \end{tabular}}
    \end{sc}
    \end{center}
\end{table*}

\begin{table*}[t]
    \caption{Objective evaluations of our base and fine-tuned models compared to \ours{}.}
    \label{tab:base_and_finetuned_results}
    \begin{center}
    \begin{sc}
    \resizebox{0.975\linewidth}{!}{%
    \begin{tabular}{lcccccccccccc}
    \toprule
     & \multicolumn{6}{c}{Short-form} & \multicolumn{6}{c}{Long-form} \\
    \cmidrule(lr){2-7}\cmidrule(lr){8-13}

    \multirow{2}{*}{Model} & & ASR & ASR & Speaker & End & & & ASR & ASR & Speaker & End & \\
                           & BLEU ($\uparrow$) & BLEU ($\uparrow$) & COMET ($\uparrow$) & Sim. ($\uparrow$) & Offset ($\downarrow$) & LAAL ($\downarrow$) & BLEU ($\uparrow$) & BLEU ($\uparrow$) & COMET ($\uparrow$) & Sim. ($\uparrow$) & Offset ($\downarrow$) & LAAL ($\downarrow$) \\
    
    \midrule
    \textbf{French} \\
    Base  % 5c94ad8f@500
        & 34.4 & 34.0 & 79.2 & 53.1 & 3.0 & 3.6
        & 31.1 & 29.7 & 43.1 & 67.5 & 3.5 & 6.2 \\
    Finetuned  % 00308e00@100
        & 34.3 & 33.9 & 78.7 & 52.9 & 3.5 & 4.0
        & 31.0 & 29.0 & 41.6 & 66.7 & 4.8 & 6.9 \\
    \ours{}  % 77f82164@110
        & 35.0 & 34.6 & 80.3 & 49.5 & 2.1 & 2.8
        & 30.6 & 28.7 & 43.7 & 61.3 & 2.3 & 6.1 \\
    
    \midrule
    \textbf{Spanish} \\
    Base
        & 34.0 & 33.7 & 80.1 & 60.2 & 3.3 & 4.1
        & 33.8 & 32.8 & 45.3 & 69.8 & 3.8 & 6.3 \\
    Finetuned
        & 33.9 & 33.7 & 80.3 & 60.2 & 3.6 & 4.2
        & 32.9 & 32.2 & 43.5 & 69.2 & 4.7 & 6.7 \\
    \ours{}
        & 33.8 & 33.9 & 80.3 & 57.0 & 2.3 & 3.1
        & 32.3 & 31.5 & 42.3 & 64.6 & 2.6 & 5.6 \\
    
    \midrule
    \textbf{Portuguese} \\
    Base
        & 33.5 & 33.2 & 78.8 & 54.2 & 3.6 & 4.2
        & 34.0 & 32.6 & 42.2 & 67.0 & 3.3 & 6.6 \\
    Finetuned
        & 34.0 & 33.9 & 78.8 & 55.5 & 3.9 & 4.3
        & 33.7 & 31.7 & 42.2 & 67.5 & 4.6 & 7.3 \\
    \ours{}
        & 33.6 & 33.6 & 78.9 & 51.4 & 2.4 & 3.0
        & 33.2 & 31.3 & 42.6 & 62.1 & 2.3 & 6.3 \\
    
    \midrule
    \textbf{German} \\
    Base
        & 28.6 & 28.4 & 83.0 & 54.8 & 2.7 & 3.6
        & 30.6 & 30.0 & 44.7 & 70.6 & 3.3 & 6.0 \\
    Finetuned
        & 28.1 & 27.9 & 82.4 & 54.6 & 2.9 & 3.7
        & 30.6 & 29.4 & 44.8 & 70.3 & 4.5 & 7.0 \\
    \ours{}
        & 28.7 & 28.6 & 82.0 & 51.5 & 1.9 & 2.8
        & 29.1 & 28.3 & 42.3 & 66.0 & 2.0 & 5.9 \\
        
    \bottomrule
    \end{tabular}}
    \end{sc}
    \end{center}
\end{table*}

\begin{table*}[t]
    \caption{Comparison between \ours{} and our model adapted for Italian on original languages with long-form evaluation.}
    \label{tab:italian_results_longform}
    \begin{center}
    \setlength{\tabcolsep}{4pt}
    \begin{sc}
    \begin{tabular}{lccccc}
    \toprule
    
    \multirow{2}{*}{Model} & BLEU         & ASR BLEU     & Speaker Sim. & End Offset     & LAAL \\
                           & ($\uparrow$) & ($\uparrow$) & ($\uparrow$) & ($\downarrow$) & ($\downarrow$) \\
    
    \midrule
    \textbf{French} \\
    \ours{}
        & 30.6 & 28.7 & 61.3 & 2.3 & 6.1 \\  % DISTIL 77f82164@110
    Italian Finetuned + RL
        & 30.6 & 29.1 & 59.8 & 3.0 & 6.2 \\  % a475c272@100
    
    \midrule
    \textbf{Spanish} \\
    \ours{}
        & 32.3 & 31.5 & 64.6 & 2.6 & 5.6 \\
    Italian Finetuned + RL
        & 31.1 & 30.3 & 62.8 & 2.6 & 6.3 \\
    
    \midrule
    \textbf{Portuguese} \\
    \ours{}
        & 33.2 & 31.3 & 62.1 & 2.3 & 6.3 \\
    Italian Finetuned + RL
        & 32.9 & 31.3 & 56.4 & 2.7 & 6.5 \\
    
    \midrule
    \textbf{German} \\
    \ours{}
        & 29.1 & 28.3 & 66.0 & 2.0 & 5.9 \\
    Italian Finetuned + RL
        & 30.5 & 28.7 & 64.2 & 2.7 & 6.4 \\
    
    \bottomrule
    \end{tabular}
    \end{sc}
    \end{center}
\end{table*}

% Cross-entropy pretraining CE curves
\begin{figure}[t]
    \centering
    \includegraphics[width=\columnwidth]{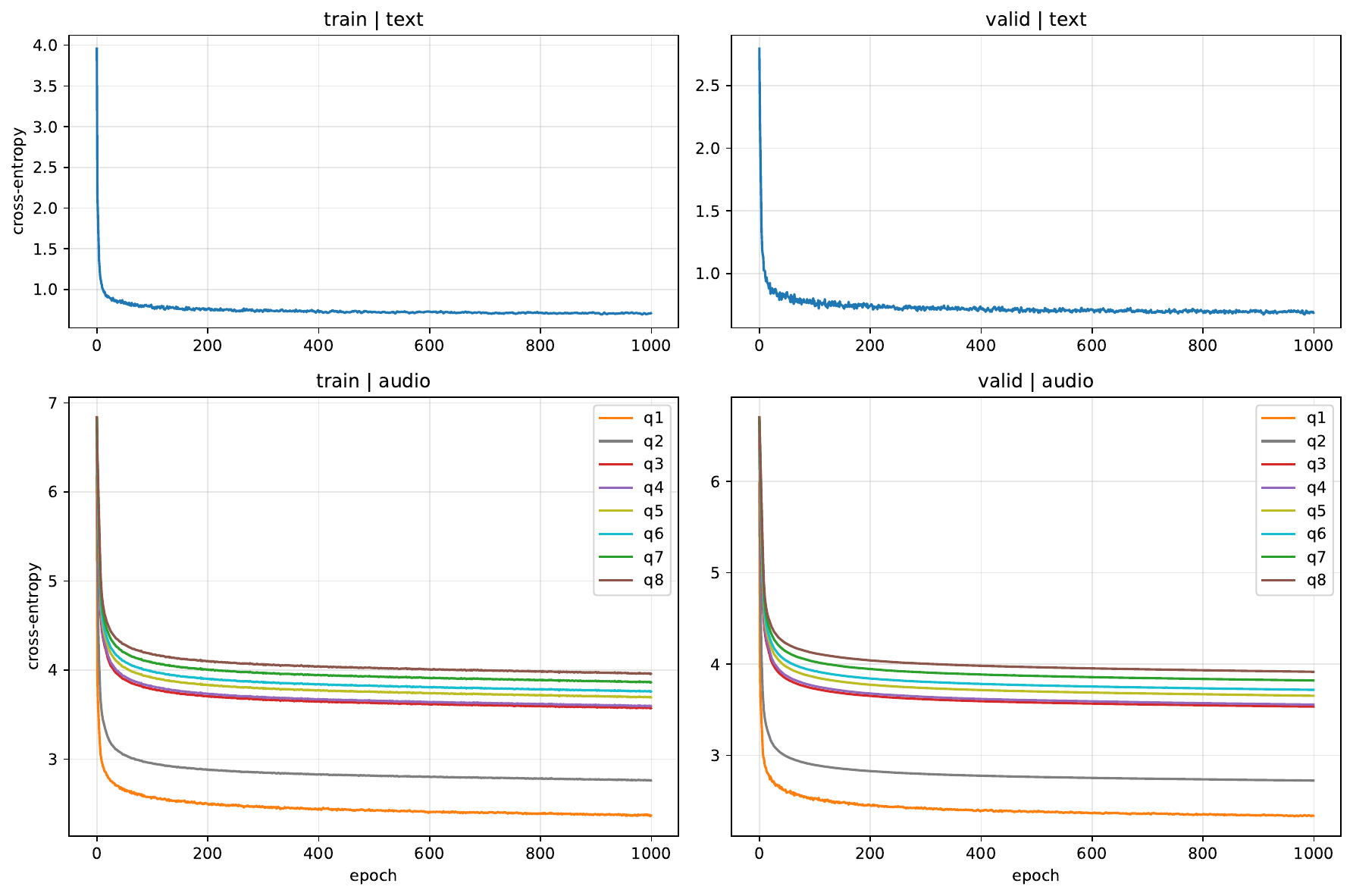}
    \caption{\textbf{Cross-entropy during audio pretraining.} Pretraining is performed with single stream audio using 8 layers of RVQ tokens.}
    \label{fig:ce_pretraining}
\end{figure}

% Cross-entropy sft CE curves
\begin{figure}[t]
    \centering
    \includegraphics[width=\columnwidth]{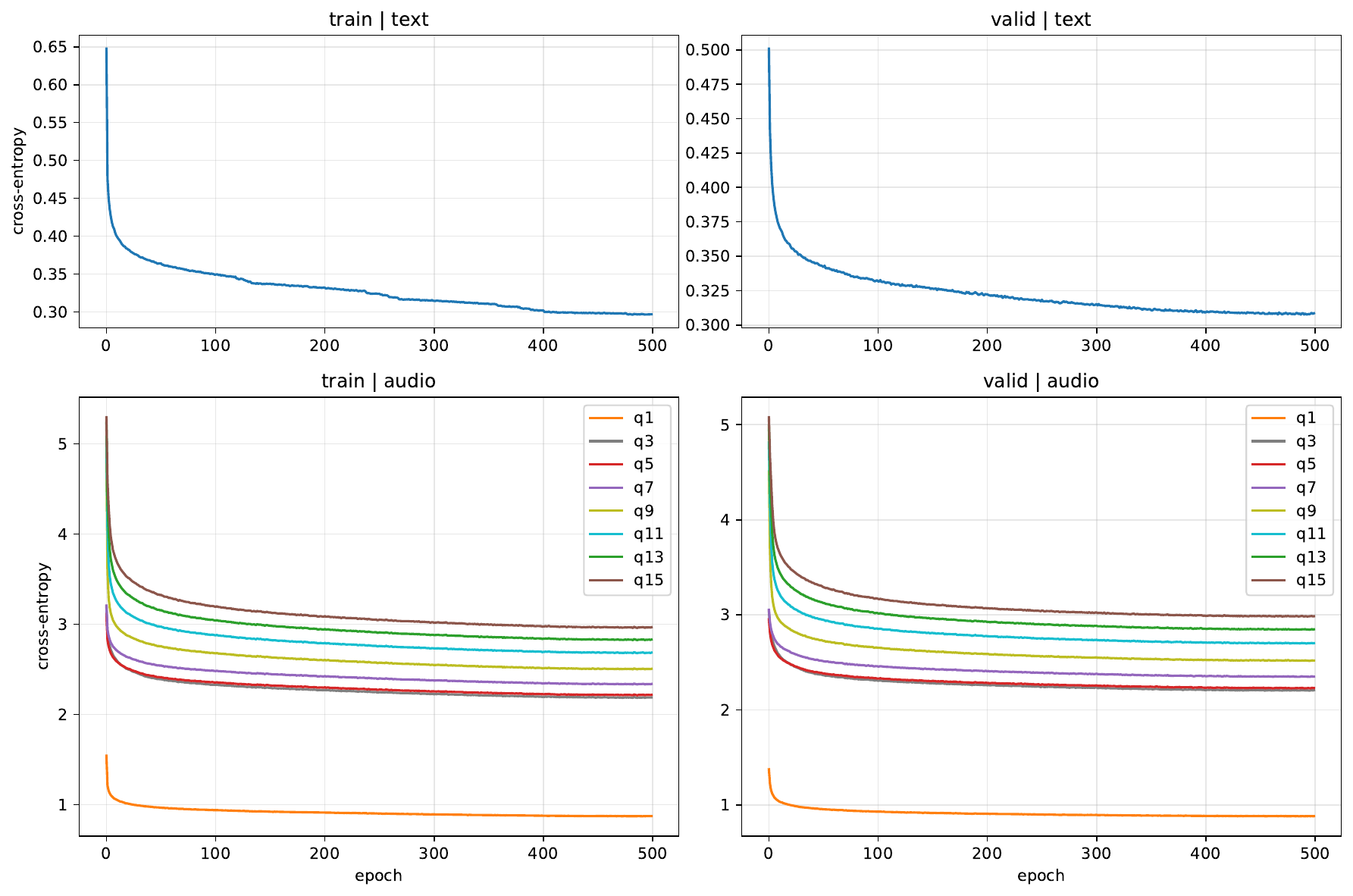}
    \caption{\textbf{Cross-entropy during coarse speech translation supervised training.} To improve readability, only the cross-entropy values of one out of every two layers of RVQ tokens from the model's audio stream are shown.}
    \label{fig:ce_sft}
\end{figure}

% Example fr outputs
\begin{figure}[t]
    \centering
    \includegraphics[width=\columnwidth]{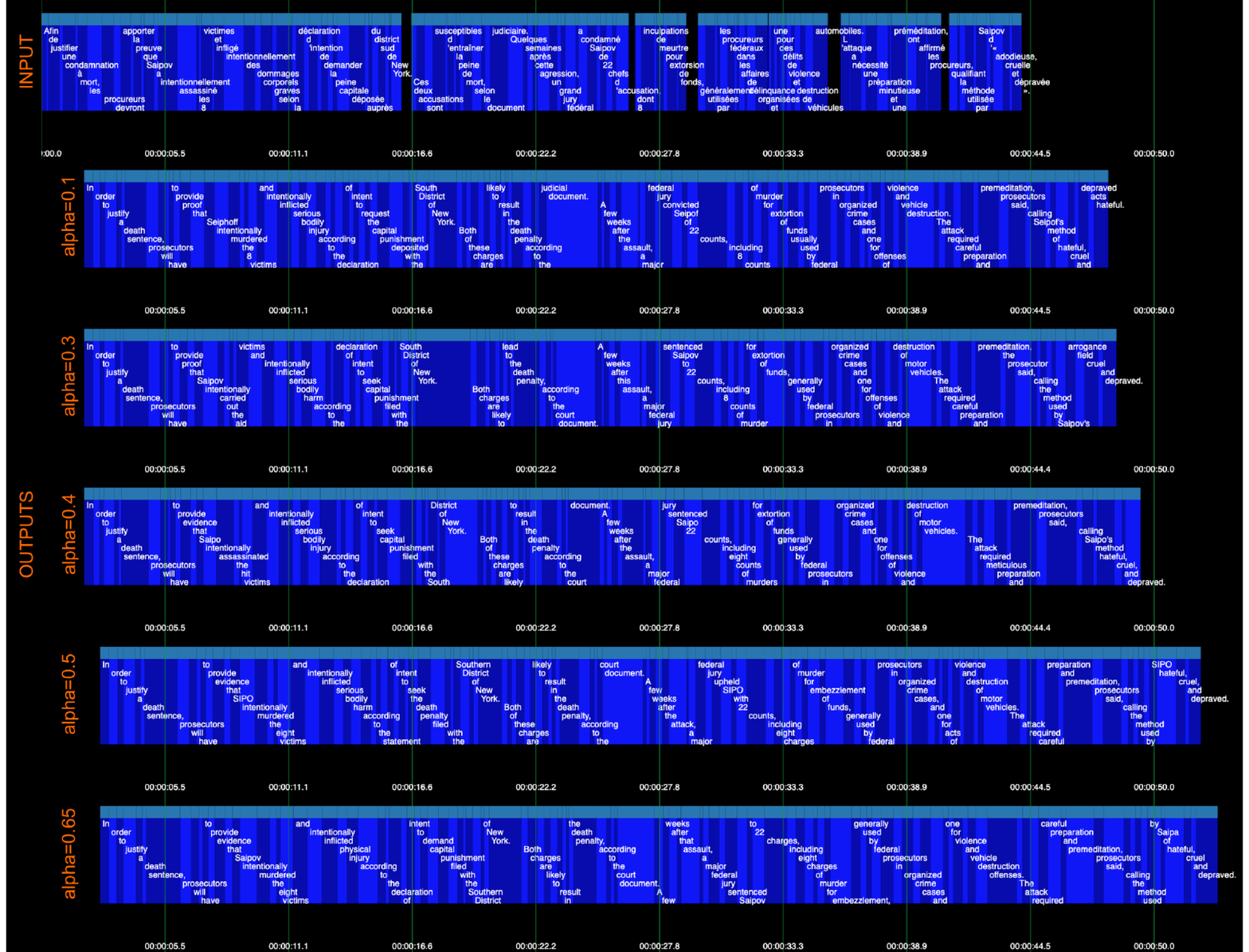}
    \caption{\textbf{Example of outputs obtained from models trained with different $\alpha$.} The input is a French sample from Audio-NTREX-4L.}
    \label{fig:fr_outputs}
\end{figure}

% Example es outputs
\begin{figure}[t]
    \centering
    \includegraphics[width=\columnwidth]{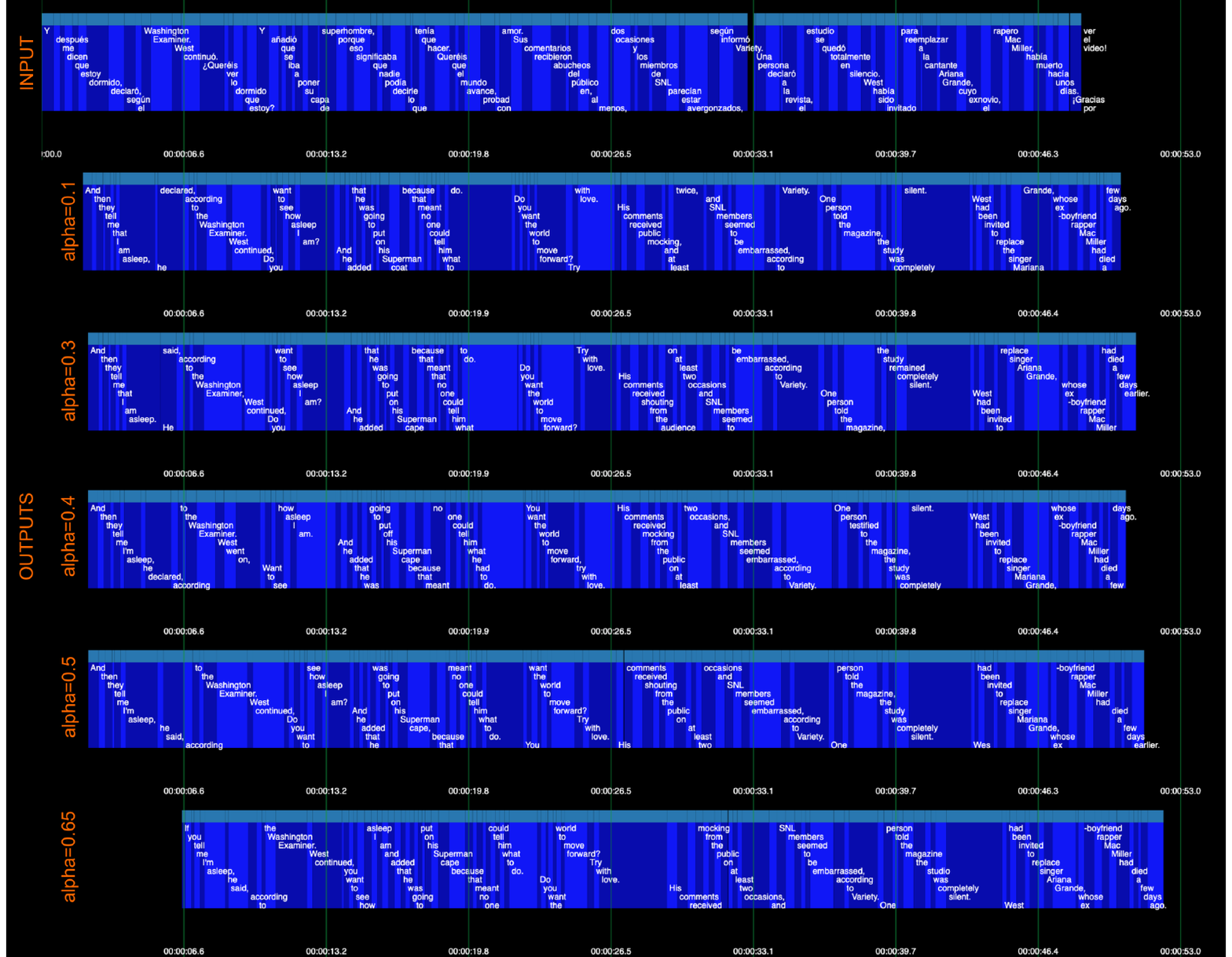}
    \caption{\textbf{Example of outputs obtained from models trained with different $\alpha$.} The input is a Spanish sample from Audio-NTREX-4L.}
    \label{fig:es_outputs}
\end{figure}

% Example pt outputs
\begin{figure}[t]
    \centering
    \includegraphics[width=\columnwidth]{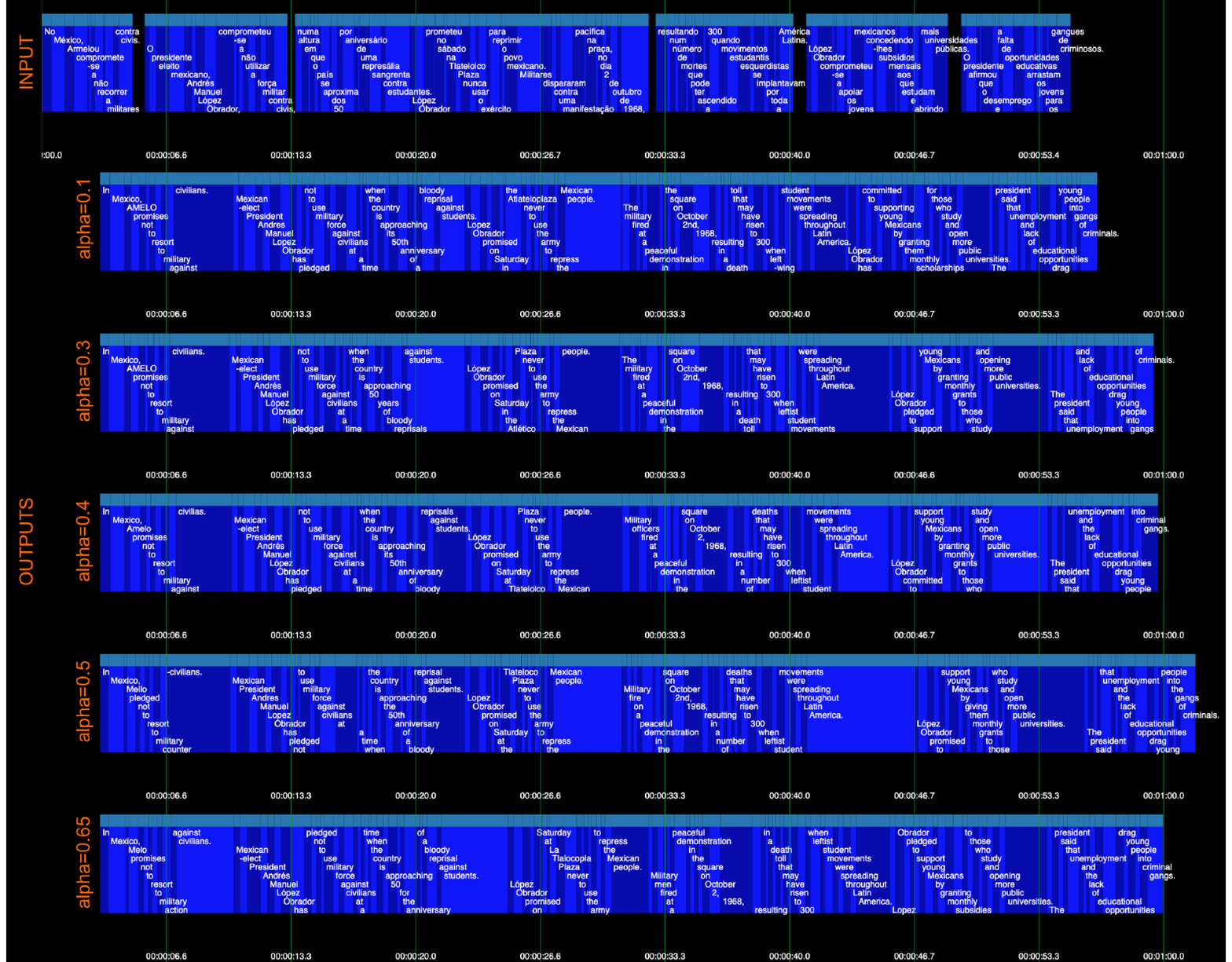}
    \caption{\textbf{Example of outputs obtained from models trained with varo $\alpha$.} The input is a Portuguese sample from Audio-NTREX-4L.}
    \label{fig:pt_outputs}
\end{figure}

% Example de outputs
\begin{figure}[t]
    \centering
    \includegraphics[width=\columnwidth]{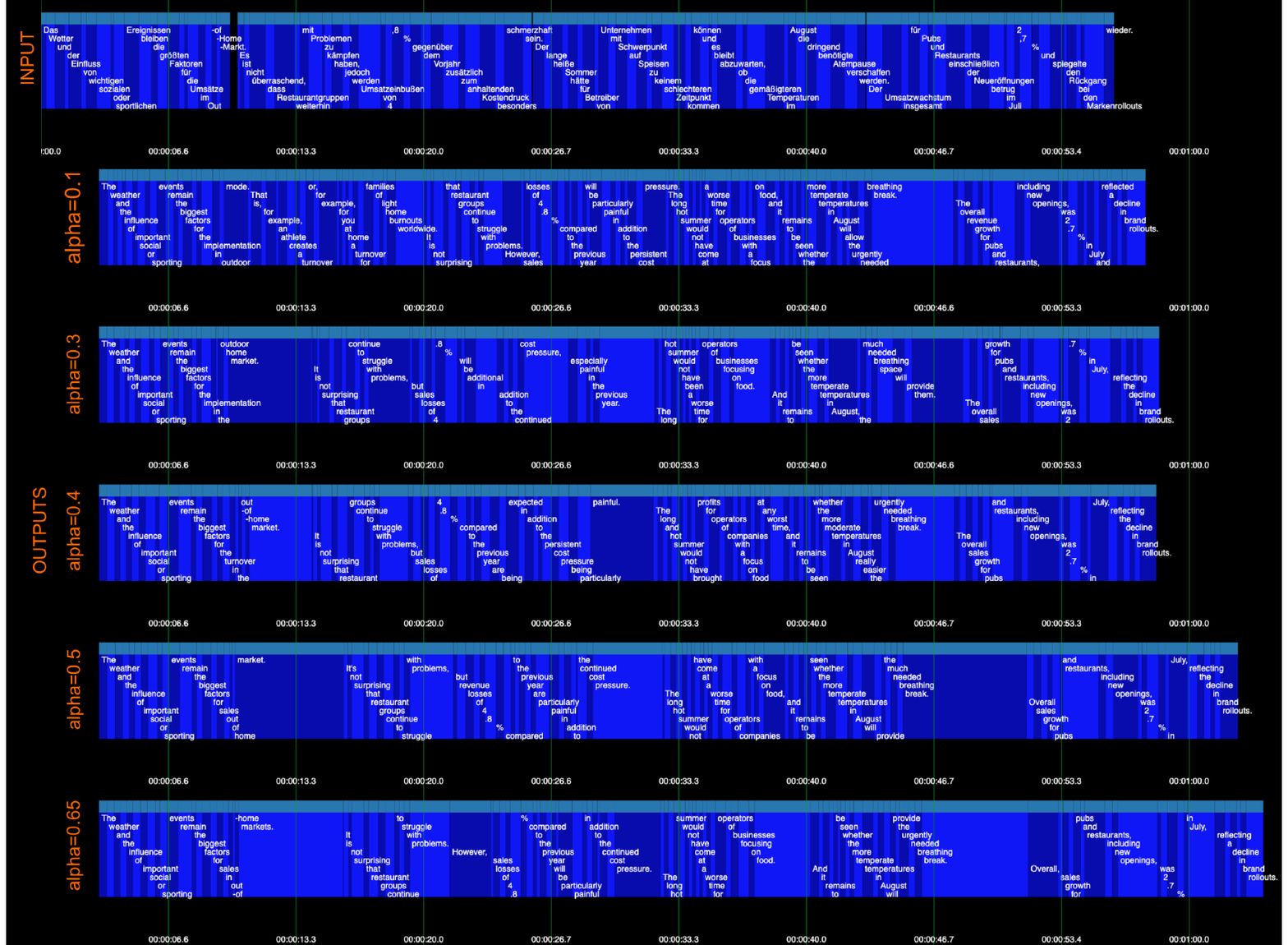}
    \caption{\textbf{Example of outputs obtained from models trained with different $\alpha$.} The input is a German sample from Audio-NTREX-4L.}
    \label{fig:de_outputs}
\end{figure}

\end{document}